\documentclass[10pt]{IEEEtran}
\usepackage{amsmath}
\usepackage{amssymb}
\usepackage{mathtools} 
\usepackage{subfig} 
\usepackage{graphicx,grffile} 
\usepackage[capitalise]{cleveref} 
\usepackage[noadjust]{cite} 
\usepackage{tabularx}
\usepackage{multirow,booktabs}
\usepackage{url}

\makeatletter
\newcases{lrdcases}
{\quad}
{$\m@th\displaystyle{##}$\hfil}
{$\m@th\displaystyle{##}$\hfil}
{\lbrace}
{} 

\newcases{lrdcases*}
{\quad}
{$\m@th\displaystyle{##}$\hfil}
{{##}\hfil}
{\lbrace}
{} 
\makeatother

\newcommand{\IGNORE}[1]{}

\newcommand{\figroot}[1]{figures/#1/fordyca}
\newcommand{\figcmproot}[1]{\figroot{#1}/d0.CRW+d0.DPO+d1.BITD_DPO+d2.BIRTD_DPO-cc-graphs}

\newcolumntype{Y}{>{\centering\arraybackslash}X}

\newcommand{\TheSwarm}{\mathbb{S}}
\newcommand{\TheSwarmSize}{\mathcal{N}}

\newcommand{\PerfMeasureHecker}[1]{P(#1)}
\newcommand{\PerfCurve}[1]{P(#1,\kappa,t)}
\newcommand{\PerfCurveSub}[2]{P_{#2}(#1,\kappa,t)}
\newcommand{\ScalabilityMetric}[1]{\mathcal{C}(#1,\kappa)}

\newcommand{\PerfLost}[1]{P_{lost}(#1,\kappa,t)}
\newcommand{\TLost}[1]{t_{lost}(#1,\kappa)}
\newcommand{\SpatialEmergenceMetric}[2]{E_{S}(#1,#2,\kappa)}
\newcommand{\TaskEmergenceMetric}[2]{E_{T}(#1,#2,\kappa)}

\newcommand{\ReactivityMetric}{R(\TheSwarmSize,\kappa)}
\newcommand{\OptimalReactivity}{R^{*}(\TheSwarmSize,\kappa)}
\newcommand{\OptimalReactivityCurve}{\PerfCurveSub{\TheSwarmSize}{R^{*}}}

\newcommand{\AdaptabilityMetric}{A(\TheSwarmSize,\kappa)}
\newcommand{\OptimalAdaptability}{A^{*}(\TheSwarmSize,\kappa)}
\newcommand{\OptimalAdaptabilityCurve}{\PerfCurveSub{\TheSwarmSize}{A^{*}}}

\newcommand{\IdealEnvConditions}{I_{ec}(t)}
\newcommand{\NonIdealEnvDeviation}{V_{dev}(t)}
\newcommand{\SARobustnessMetric}{B_{sa}(\TheSwarmSize,\kappa)}
\newcommand{\OptimalSARobustness}{B_{sa}^{*}(\TheSwarmSize,\kappa)}
\newcommand{\OptimalSARobustnessCurve}{\PerfCurveSub{\TheSwarmSize}{{B_{sa}}^{*}}}

\newcommand{\PDRobustnessMetric}{B_{pd}(\kappa)}
\newcommand{\OptimalPDRobustness}{B_{pd}^{*}(\kappa)}

\newcommand{\TaskedSwarm}{{S}}
\newcommand{\UntaskedSwarm}{\bar{S}}
\newcommand{\AssignedTask}{\mathcal{T}}
\newcommand{\TimeInTaskedSwarm}{T_{\TaskedSwarm}}
\newcommand{\TimeInTaskedSwarmIdeal}{T_{\TaskedSwarm_{ideal}}}
\newcommand{\TimeInUntaskedSwarm}{T_{\UntaskedSwarm}}

\newcommand{\TotalTime}{T}

\newcommand{\TaskedSwarmSize}[1]{N(#1)}
\newcommand{\TaskedSwarmSizeMin}{N_{min}}
\newcommand{\BirthQueue}{Q_{b}}
\newcommand{\DeathQueue}{Q_{d}}
\newcommand{\RepairQueue}{Q_{r}}

\newcommand{\UntaskedSwarmQueue}{Q_{\UntaskedSwarm}}
\newcommand{\UntaskedSwarmQueueLambda}{\lambda_{d} + \lambda_{bd}}
\newcommand{\UntaskedSwarmQueueMu}{\mu_{b} + \mu_{bd}}

\title{Improved Swarm Engineering: Aligning Intuition and Analysis}

\author{John Harwell,~\IEEEmembership{Member,~IEEE,}
and Maria Gini,~\IEEEmembership{Fellow,~IEEE}%
\thanks{Work supported in part by Amazon Robotics, the MnDRIVE RSAM initiative at the University of Minnesota, the Minnesota Supercomputing Institute,
and the UMII Graduate Assistantship from the University of Minnesota
}%
\thanks{J, Harwell and M. Gini are with the Department of Computer Science
and Engineering, University of Minnesota, Minneapolis, MN 55455, USA
(e-mail: harwe006@umn.edu, gini@umn.edu)}
}
\begin{document}

\maketitle

\begin{abstract}
  We present a set of metrics intended to supplement designer intuitions when
  designing swarm-robotic systems, increase accuracy in extrapolating swarm
  behavior from algorithmic descriptions and small test experiments, and lead to
  faster and less costly design cycles. We build on previous works studying
  self-organizing behaviors in autonomous systems to derive a metric for swarm
  emergent self-organization.  We utilize techniques from high performance
  computing, time series analysis, and queueing theory to derive metrics for
  swarm scalability, flexibility to changing external environments, and
  robustness to internal system stimuli such as sensor and actuator noise and
  robot failures.

  We demonstrate the utility of our metrics by analyzing four different control
  algorithms in two scenarios: an indoor warehouse object transport scenario
  with static objects and a spatially unconstrained outdoor search and rescue
  scenario with moving objects. In the spatially constrained warehouse scenario,
  efficient use of space is key to success so algorithms that use mechanisms for
  traffic regulation and congestion reduction are the most appropriate.  In the
  search and rescue scenario, the same will happen with algorithms which can
  cope well with object motion through dynamic task allocation and randomized
  search trajectories. We show that our intuitions about comparative algorithm
  performance are well supported by the quantitative results obtained using our
  metrics, and that our metrics can be synergistically used together to predict
  collective behaviors based on previous results in some cases.
\end{abstract}


%
\section{Introduction}\label{sec:intro}
Swarm Robotics (SR) is the study of the coordination of large numbers of simple
robots~\cite{Sahin2005}. SR systems can be homogeneous (single robot type and
identical control software) or heterogeneous (multiple robot types and/or
control software)~\cite{Dorigo2013,Rizk2019,Ramachandran2020}. The main
differentiating factors between SR systems and multi-agent robotics systems stem
from the mechanisms on which SR systems are based.  Historically, these were
principles of biological mimicry or problem solving techniques inspired by
natural systems of agents such as bees, ants, and
termites~\cite{Labella2006}.

The duality between SR and natural systems enables effective parallels to be
drawn with many naturally occurring problems, such as foraging, collective
transport of heavy objects, environmental monitoring, object sorting, hazardous
material cleanup, self-assembly, exploration, and collective decision
making~\cite{Hecker2015,Kumar2003,CarrilloZapata2020}.  As a result, SR systems
are especially suited for complex tasks in dynamic environments where robustness
and flexibility are key to success.  

Moving beyond strictly biomimetic design, many modern SR systems have retained
the desirable properties of natural systems while employing
elements of more conventional multi-robot system
design~\cite{Castello2016}. Those properties are:

\textbf{Emergent Self-Organization}. \emph{Self-organization} is the appearance
of structure within the swarm, which can be spatial, temporal, or functional. It
arises at the \emph{collective} level due to inter-robot interactions and robot
decisions at the \emph{individual} level~\cite{Winfield2005a,Galstyan2005}. Once
established, self-organization is generally resistant to external stimuli, and
this resistance is crucial to solving complex problems with only simple
agents~\cite{Hunt2020,DeWolf2005}.  Self-organization is related to, but
distinct from, the concept of \emph{emergence}, which is the set difference
between individual and collective swarm behavior, where the swarm behavior
arises both from inter-robot interactions and individual robot
decisions~\cite{Szabo2014,DeWolf2005}.  Emergent and self-organizing behaviors
are often both present in SR systems as robots collectively find solutions to a
problem that they cannot solve alone~\cite{Cotsaftis2009,Hunt2020}. We term this
dichotomy \emph{emergent self-organization}, i.e, a two-way link between
collective and individual level behaviors.

\textbf{Scalability}. Scalable systems are able to maintain efficiency and
effectiveness as system size grows. SR systems, like natural systems, can
achieve scalability to hundreds or thousands of agents through decentralized
control~\cite{Matthey2009}; other methods of maintaining effectiveness at larger
scales include local agent communication~\cite{Agassounon2001,Lerman2006} and
heterogeneous robots or robot roles~\cite{Lu2020,Harwell2019a}.  Scalability can
arise directly from the design of the swarm control algorithm, but more often as
a cumulative result of emergent self-organizing behaviors which are themselves
decentralized and local~\cite{DeWolf2005}.

\textbf{Flexibility}. Flexible systems are able to modify their collective
behavior in response to unknown stimuli in the external environment, e.g.,
changing weather conditions~\cite{Harwell2019a,Just2017,Hunt2020}. Individual
agents make decisions based on locally available information from neighbors and
their own limited sensor data. This produces a spatially distributed response
which arises from robot decisions as the swarm reacts and adapts to the
environment. SR systems, like natural systems, can achieve flexibility in a
variety of ways: pheromone trails, site fidelity, localized communication, and
task allocation strategies~\cite{Just2017,Harwell2019a}. Some analytical methods,
such as task allocation strategies, explicitly encode mechanisms for the swarm
to collectively attempt to mitigate adversity and exploit beneficial changes in
dynamic environmental conditions~\cite{Just2017,Winfield2008}. Other methods,
such as pheromone trails, rely on the plasticity of self-organizing behaviors
and the development of emergent behaviors to achieve flexibility. The
flexibility of such methods can therefore be coupled to, but is still distinct
from, emergent and self-organizing behaviors~\cite{DeWolf2005}.

\textbf{Robustness}. Robust systems are able to modify their collective behavior
in response to internal, as opposed to environmental, stimuli.  Such
stimuli can include sensor and actuator noise, changes in system size due to the
introduction of new robots, and robot failures.
Robustness is therefore an emergent property of systems which demonstrate
resilience to the effects of internal stimuli on individual
robots~\cite{DeWolf2005} (e.g., losing a single robot or set of robots minimally
perturbs the collective behavior of the swarm).  Robustness is crucial for
crossing the simulation-reality gap~\cite{Hecker2015,Francesca2014}.  Some SR
systems handle sensor and actuator noise
analytically~\cite{Dallalibera2011,Claudi2014,Turgut2008}, and can provide
theoretical guarantees of robustness.  Other systems rely on emergent
behaviors~\cite{Harwell2020a} and do not provide theoretical guarantees; in such
cases, robustness is coupled to, but again distinct from, emergent
self-organization. 

The duality between designing swarm control algorithms directly to be scalable
and self-organizing, while incorporating emergent behaviors for flexibility and
robustness, can be leveraged in the design of practical SR systems.
First formally discussed by~\cite{Winfield2005}, \emph{swarm engineering} seeks
to design systems which are based on rigorous mathematical methods but which
also possess the desirable system properties discussed
above~\cite{Brambilla2013a}.  Each of these desirable properties can be designed
for by answering the following design questions:
\begin{enumerate}
\item {Does the solution show emergent self-organization, indicating collective
    intelligence and its potential to be used in variants of the given problem?
  }
\item{Does the solution scale to the current and future needs of the modeled
    scenario?}
\item {Can the solution flexibly handle unknown environments or those with
    fluctuating conditions?}

\item {Is the solution robust to sensor/actuator noise and population size
    fluctuations? 
  }
\end{enumerate}
In order to answer these questions we first need a method for quantifying each
of the desirable system properties with numerical calculations; to the best of
our knowledge, such a methodology does not exist. The main contribution of this
paper is the establishment of such a methodology and several demonstrations of
its utility. Specifically, we propose:


\begin{enumerate}
\item \textbf{Metrics for SR systems properties}.  We present metrics for swarm
  emergent self-organization, scalability, flexibility, and robustness as
  analysis tools to assist with answering our design questions
  (\cref{sec:meas-emergence,sec:meas-scalability,sec:meas-flexibility,sec:meas-robustness}). Our
  approach is domain agnostic and serves as a starting point to develop
  application-specific variants. Furthermore, we establish the groundwork for
  more comprehensive theories of swarm behavior, because measuring system
  properties precisely is a necessary precursor to developing theories about
  what elements of a swarm control algorithm give rise to observed behaviors.
\item \textbf{Realistic scenario modeling.}  We apply our metrics to two
  real-world problems: indoor warehouse object transport (\cref{sec:sc1}) and
  outdoor search and rescue (\cref{sec:sc2}).  Through application to these
  complex real-world scenarios we expand the range of characteristics affecting
  swarm behavior which can be meaningfully studied in simulation. For example,
  the ability to precisely measure how different levels of sensor and actuator
  noise affect swarm behavior allows us to incorporate noise-generating elements
  of real-world problems into our scenario model and better study their
  effects. Without such methodology, those effects can only be studied
  qualitatively, which is not as useful.
\end{enumerate}
%
%
A preliminary version of this paper was published in the IJCAI
conference~\cite{Harwell2019a}. It provided metrics for scabalility,
emergent self-organization, and flexibility, and evaluated them in an indoor
warehouse setting. This paper extends previous work in the following ways:
\begin{enumerate}
\item {Refined metrics for emergent self-organization, scalability,  and
    flexibility, and added metrics for robustness.}
\item {Applying our metrics to a search and rescue scenario, in addition to an
    indoor warehouse scenario.}
\item {Adding two methods that use a different way of doing task allocation to
    the set of swarm control algorithms which are used for comparison in each
    scenario. }
\item {More thorough discussion of recent related work in swarm engineering and
    our metrics.  }
\end{enumerate}
The rest of the paper is organized as follows.~\cref{sec:related-work} discusses
related works in swarm engineering. We discuss swarm emergent self-organization,
scalability, flexibility, and robustness, and derive metrics for each
in~\cref{sec:meas-emergence,sec:meas-scalability,sec:meas-flexibility,sec:meas-robustness}.
\cref{sec:sc-overview} describes our two real-world problems along with design
constraints and candidate solutions. In \cref{sec:sc1,sec:sc2} we apply our
metrics to those two problems. We conclude with a discussion
in~\cref{sec:discussion} of common themes between the two scenarios.


\section{Related Work}\label{sec:related-work}

Previous work in swarm engineering can be broadly classified into two
overlapping categories: automatic robot controller
synthesis~\cite{Francesca2014} which may guarantee collective behaviors and
system
properties~\cite{Winfield2005,Campo2007,Turgut2008,Correll2008,Moarref2018},
and the use of mathematical techniques to analyze and predict the collective
behavior of swarms from characteristics of individual robot
controllers~\cite{Talamali2020,Lerman2006,Berman2007}.

A method for optimal foraging, as defined from a biological perspective, was
presented in~\cite{Talamali2020} based on pheromone trails. Differential
equations were used to model the collective behavior of reactive
robots~\cite{Lerman2001} and robots with memory which perform task
allocation~\cite{Lerman2006}. Evolutionary techniques have also been used, and
the resulting controllers shown to be capable of successfully crossing the
simulation-reality gap~\cite{Francesca2014,Ligot2020}; other
works~\cite{Hogg2020} have evolved swarm supervisors (e.g., replacements for
human supervisors).

Controller synthesis via combined Lyapunov functions and control theoretic
techniques can provide guarantees that the resulting swarm is ``safe'' (i.e.,
stay in a desired set of states) in problems such as area patrol, autonomous
driving, and robot
walking~\cite{Panagou2020,Glotfelter2019,Ames2021}. Similarly, ensemble dynamics
filtering has been used to provably shape swarm dynamics during search and
rescue~\cite{Hsieh2013}. Temporal logic has been used as an alternative basis
for controller synthesis in scenarios where the results of robot actions are not
immediately observable~\cite{Winfield2005,Moarref2018} (e.g., robot motion
between spatial locations), for identifying unrealizable controllers and
analyzing synthesizer output~\cite{Baumeister2020}, and automated suggestion to
make unrealizable controllers realizable~\cite{Pacheck2020}. For a review of
recent works and issues in controller synthesis, see~\cite{Birattari2019}.

Some previous works have considered swarm behavior in real-world settings with
simulated or real robots in scenarios with non-ideal characteristics; that is,
they systematically study how swarm behavior scales, exhibits flexibility, or
exhibits robustness to sensor and actuator noise when ideality assumptions
commonly employed during algorithm development are relaxed. Notable
simulation-based examples include~\cite{Harwell2019a,Bjerknes2013} and examples
with real robots include~\cite{Rubenstein2014}.

From a swarm engineering perspective, consider which of the following algorithms
would have greater utility: a heuristic control algorithm with no provable
guarantees but good performance on many problems of interest, or a rigorously
derived control algorithm with many theoretical guarantees but relatively poor
performance on problems of interest?  Depending on the application parameters,
stakeholders, and other factors, either algorithm may be considered the
``best.''; this echoes the concept of ecological rationality, in which the
\emph{actual} performance of an organism in an environment is more important
than its \emph{potential} performance/how well adapted it is to an environment
on paper.

Historically speaking, the ``best'' algorithm is selected through iterative
refinement: testing, algorithmic study, etc.  The quality of the solution
depends strongly on the experience of the designer~\cite{Francesca2014}. The
number of design iterations can be reduced through (1) more accurate problem
modeling, or (2) quantitative metrics for system properties that can be used to
predict performance on variances of the problem.
Previous
work~(\cite{Castello2016,Harwell2018}) in swarm engineering on algorithm
selection for a target application only implicitly considers the swarm
properties through raw performance comparisons; a methodology capable of (1) and
(2) was presented in~\cite{Harwell2019a} for some swarm properties.

\section{Metric Summaries and Preliminaries}\label{sec:metrics-preliminaries}
In the sections that follow we define metric axes and derive metrics for swarm
emergent self-organization, scalability, flexibility, and robustness
in~\cref{sec:meas-scalability,sec:meas-emergence,sec:meas-flexibility,sec:meas-robustness}. We
have defined our axes and metrics to be as general as possible and applicable to
almost all domains within multi-robot systems; more precise axes specific to a
given domain may provide deeper insight than our definition at the expense of
limited wider applicability.

The notation we use is summarized in~\cref{tab:notation-summary}.  A taxonomy of
SR properties with our metric axes is in~\cref{tab:sr-taxonomy}, and a top-level
summary of our metrics is in~\cref{tab:measures-summary}.

\begin{table}[t]
  \centering\scriptsize
  \begin{tabularx}{\linewidth}{ p{1.5cm} X}
    {Quantity} &  {Description}  \\

    \toprule

    $\TheSwarm$ & A swarm of $\TheSwarmSize$ homogeneous robots, each running $\kappa$.  \\[1ex]

    $\kappa$ & The robot control algorithm present on each robot in $\TheSwarm$. \\[1ex]

    $\TheSwarmSize$ & The number of robots in $\TheSwarm$. \\[1ex]

    $\TheSwarmSize_{1}, \TheSwarmSize_{2}$ & Two specific swarm sizes, with
    $\TheSwarmSize_{1} < \TheSwarmSize_{2}$. \\[1ex]

    $t, \TotalTime$ & A single time step of execution/the total execution time of
    $\TheSwarm$. \\[1ex]

    $\TaskedSwarm$, $\UntaskedSwarm$ & The sub-swarm
    $\TaskedSwarm\subseteq\TheSwarm$ ($\UntaskedSwarm\subseteq\TheSwarm$) which
    is (is not) working on an assigned task $\AssignedTask$ of interest. \\[1ex]

    $\AssignedTask$ & A task of interest which $\TaskedSwarm$ is working on at $t$.
    $\TheSwarm$ may be working on multiple tasks simultaneously. \\[1ex]

    $\PerfCurve{\TheSwarmSize}$ & The temporal performance curve a swarm
    $\TheSwarm$ traces out over time, measurable at each $t$ under
    $\NonIdealEnvDeviation$. \\[1ex]

    $\PerfCurveSub{\TheSwarmSize}{ideal}$ & The temporal performance curve a
    swarm $\TheSwarm$ traces out over time, measurable at each $t$ under
    $\IdealEnvConditions$.  \\[1ex]

    $\TLost{\TheSwarmSize}$ &Fraction of time lost to inter-robot
    interference within $\TheSwarm$. \\[1ex]

    $\PerfLost{\TheSwarmSize}$ & Fraction of performance lost to inter-robot
    interference within $\TheSwarm$. \\[1ex]

    $\IdealEnvConditions$ & Continuous one-dimensional signal representing the
    ideal
    environmental conditions or the conditions that $\kappa$ was developed in.  \\[1ex]

    $\NonIdealEnvDeviation$ & Continuous one-dimensional signal representing the
    signed waveform of an adverse deviation from $\IdealEnvConditions$, where
    positive values indicate increased adversity in environmental conditions in
    comparison with $\IdealEnvConditions$, and negative values
    indicate decreased adversity. \\[1ex]

    $\OptimalReactivity$ & Optimal reactivity, defined as
    $c_t\PerfCurveSub{\TheSwarmSize}{ideal}$, where $c_t$ is a
    non-zero constant per $t$ dependent on the value of
    $\NonIdealEnvDeviation$. \\[1ex]

    $\OptimalAdaptability$ & Optimal adaptability curve, where the Dynamic Time Warp
    $DTW(\PerfCurveSub{\TheSwarmSize}{ideal},\PerfCurve{\TheSwarmSize})=0$,
    regardless of $\NonIdealEnvDeviation$. \\[1ex]

    $\OptimalSARobustness$ & Optimal sensor and actuator noise robustness,
    where
    $DTW(\PerfCurveSub{\TheSwarmSize}{ideal},\PerfCurve{\TheSwarmSize})=0$.  \\[1ex]

    $\OptimalPDRobustness$ & Optimal population dynamics robustness, where
    $\PerfCurve{\TheSwarmSize}\ge\PerfCurveSub{\TheSwarmSize}{ideal}$ for all $t$,
    despite fluctuating $\TheSwarmSize$.  \\[1ex]
    \bottomrule
  \end{tabularx}\caption{\footnotesize{ Notational definitions used throughout
      our derivations.}}\label{tab:notation-summary}
\end{table}
\begin{table}[ht]
  \centering
  \begin{tabular}{p{2.4cm} @{} p{2.8cm} @{~} p{2.8cm}}\hline
    {Swarm Property} &  {Property Axis 1}  & {Property Axis 2}  \\
    \hline
    Emergent \\ Self-Organization &  Spatial & Task-based \\[1ex]
    Scalability &  Degree of cooperation & N/A \\[1ex]
    Flexibility &  Reactivity & Adaptability \\[2ex]
    Robustness  &  Sensor \& actuator noise & Population dynamics  \\[1ex]
    \hline
  \end{tabular}\caption{\footnotesize{Taxonomy of the SR system properties and the axes along
      which we have derived quantitative metrics for each property. }}\label{tab:sr-taxonomy}
\end{table}
\begin{table*}[t]
  \centering\scriptsize
  \begin{tabularx}{\linewidth}{p{1.75cm} p{1.5cm} X}
    Swarm Property & Notation & Summary and Intuition\\
    \toprule

    \multirow{2}{*}{\begin{minipage}[t]{\linewidth}Emergent
        Self-Organization\end{minipage}} &
    $\SpatialEmergenceMetric{\TheSwarmSize_{1}}{\TheSwarmSize_{2}}$ & The
    \emph{Spatial} emergent self-organization of $\TheSwarm$. Computed via the
    degree of \emph{sub}-linearity of inter-robot interference with
    $\TheSwarmSize_{1} < \TheSwarmSize_{2} $ swarm sizes. \\[1ex]

    & $\TaskEmergenceMetric{\TheSwarmSize_{1}}{\TheSwarmSize_{2}}$ & The
    \emph{Task} emergent self-organization of $\TheSwarm$. Computed via the
    degree of \emph{super}-linearity of performance gains with
    $\TheSwarmSize_{1} < \TheSwarmSize_{2}$ swarm sizes. \\[1ex]

    \midrule Scalability &
    $\ScalabilityMetric{\TheSwarmSize_{1},\TheSwarmSize_{2}}$ & The fraction of
    the performance of $\TheSwarm$ due to (parallel) cooperation, opposed to
    (serial) independent action. Computed via a modified version of the
    Karp-Flatt metric for analyzing potential
    speedups in high
    performance computing systems. \\[1ex]

    \midrule
    \multirow{2}{*}{Flexibility} &

    $\ReactivityMetric$ & The proportional \emph{Reactivity} of $\TheSwarm$ to
    fluctuating changes in environmental conditions over time: if conditions
    improve, performance improves, if conditions worsen, performance
    worsens. Computed via the Dynamic Time Warp (DTW) distance of
    $\PerfCurve{\TheSwarmSize}$ from the optimal reactivity curve
    $\OptimalReactivityCurve$. \\[1ex]

    & $\AdaptabilityMetric$ & The \emph{Adaptability} of $\TheSwarm$ to
    fluctuating changes in environmental conditions: 
    regardless of how
    conditions change, performance remains the same. Computed via the
    Dynamic Time Warp (DTW) distance of $\PerfCurve{\TheSwarmSize}$ from the
    optimal adaptability curve
    $\OptimalAdaptabilityCurve$. \\[1ex]

    \midrule
    \multirow{2}{*}{Robustness} &

    $\SARobustnessMetric$ & The robustness of $\TheSwarm$ to noisy sensors and
    actuators. Regardless of how noisy sensors and actuators are, the
    performance of $\TheSwarm$ should remain the same as with noise-free sensors
    and actuators. Computed via the \emph{Dynamic Time Warp} (DTW) distance of
    $\PerfCurve{\TheSwarmSize}$ from
    the ideal performance curve $\PerfCurveSub{\TheSwarmSize}{ideal}$. \\[1ex]

    & $\PDRobustnessMetric$ & The robustness of $\TheSwarm$ to fluctuating
    population sizes ($\TaskedSwarmSize{t}$) over time as a result of permanent
    or temporary robot failures and task reallocations. Computed via the degree
    of \emph{sub}-linear of performance decreases as the swarm population increases its variability.  \\[1ex]
    \bottomrule
  \end{tabularx}\caption{\footnotesize{Summary of our derived metrics,
      their intuition, and calculation.}}\label{tab:measures-summary}
\end{table*}

$\PerfCurve{\TheSwarmSize}$ is an arbitrary performance measurement of some aspect of
the behavior of $\TheSwarm$:
time to complete a certain number of tasks, task completion rate, etc. This
formulation of swarm performance allows us to simultaneously analyze (1) cumulative
performance by summing across all $t\in{T}$ (scalability and emergent
self-organization), and (2) temporally varying performance curves via pairwise point
comparison for each $t\in{\TotalTime}$ (flexibility and robustness). We proceed with
the following assumptions:
\begin{enumerate}
\item {$\PerfCurve{\TheSwarmSize} \ge 0 $ for all $t$. }
\item {If $\kappa_1$ is considered to be better than $\kappa_2$ at time $t$,
    then $P(\TheSwarmSize,\kappa_1,t) >
    P(\TheSwarmSize,\kappa_2,t)$. Performance measures where smaller values are
    better (e.g., average time to task completion) can be made to satisfy this
    assumption by taking their reciprocal at each $t$.}
\end{enumerate}
%

%
\section{Swarm Emergent Self-Organization}\label{sec:meas-emergence}

\subsection{Background}\label{ssec:meas-emergence-bg}

\begin{figure}[t]
  \centering
  \includegraphics[width=.95\linewidth]{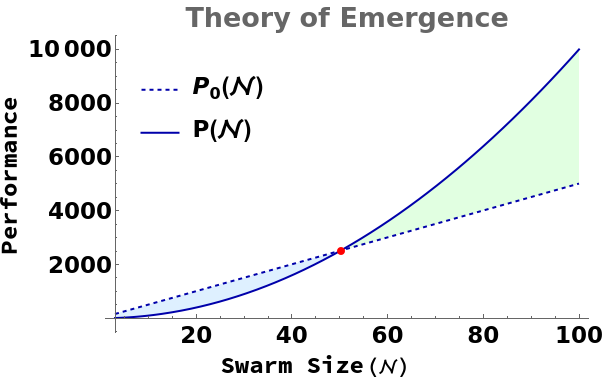}
  \caption{\label{fig:emergence-theory} \footnotesize{Visual representation of
      our definition of emergence, where
      $P(\TheSwarmSize) - P_0(\TheSwarmSize) > 0$. $P_0(\TheSwarmSize)$ is the
      performance attained by $\TheSwarmSize$ independent robots, and
      $P(\TheSwarmSize)$ is the performance achieved by $\TheSwarmSize$
      interacting robots. In this example, emergence occurs when
      $\TheSwarmSize \ge 50$ (light green shaded region).}}
\end{figure}
Within SR, no widely accepted theory of self-organizing systems
exists
~\cite{Hamann2008,Galstyan2005}, in part because emergent, self-organizing
swarms cannot be easily understood by studying individual parts in
isolation~\cite{DeWolf2005}. In SR systems, interactions between components can
overtake outside interactions and give rise to new emergent behaviors not
predictable from component study; that is, directly from the swarm control
algorithm~\cite{Cotsaftis2009,George2005,Hunt2020,DeWolf2005}. However,
self-organization can be related to the latency of information propagation
through a swarm and swarm sparseness~\cite{Tarapore2020}, and is often
synergistically coupled to emergent behaviors~\cite{DeWolf2005}.

While many papers show evidence that their algorithms exhibit emergent
behavior~(\cite{Frison2010,Harwell2018,Matthey2009}), or provide a framework for
logical reasoning about emergent
properties~(\cite{Winfield2005a,Li2016,Beni2005}), a general theory of emergence
remains elusive. Nevertheless, some common preconditions for self-organizing
emergent behaviors have been
observed
~\cite{Bonabeau1999,DeWolf2005}. Let $\TheSwarm$ be a swarm of $\TheSwarmSize$
robots, then the preconditions are:
\begin{enumerate}
\item {\emph{Positive feedback} via heuristics which promote organization at the
    collective level, such as reactive recruitment and reinforcement. In
    constrained physical environments it can lead to emergent spatial
    self-organization~\cite{Correll2006}.}

\item {\emph{Negative feedback} via pheromone evaporation, interference, etc.,
    stabilizes the behavior of $\TheSwarm$
~\cite{Payton2001,Harwell2018}.}

\item {\emph{Random fluctuations} via random walks, random task switching, etc.,
    are necessary for collective creativity and problem solving, though they can
    also lead to sub-optimal collective solutions if not moderated by other
    properties.}

\item {\emph{Multiple interactions} between robots.  Robots use information from
    other robots, possibly through joint environment manipulation, to choose how
    to act. More interactions can lead to more informed decisions.}
\end{enumerate}
Early analyses of emergent behavior posited that a swarm of $\TheSwarmSize$
units only exhibits emergent behavior when $\TheSwarmSize$ exceeds a critical
size $\TheSwarmSize_{c}$~\cite{Beni1993}. Below this threshold, swarm attempts
at self organization can negatively affect performance
(\cref{fig:emergence-theory}, light blue shaded region). A more generalized
definition defines emergence as the \emph{positive difference}
$P(\TheSwarmSize) - P_0(\TheSwarmSize)$~\cite{Sugawara1997}
(\cref{fig:emergence-theory}, shaded region), where $P(\TheSwarmSize)$ is the
amount of performance achieved by a swarm of $\TheSwarmSize$ 
robots, and $P_0(\TheSwarmSize)$ is the amount of performance achieved by an
equivalent swarm of independent agents; this intuition is also mentioned
in~\cite{Lerman2001,Hunt2020,DeWolf2005}. In essence, emergent systems exhibit
non-linearities in behavior; self-organizing systems also exhibit
non-linearities as they acquire and maintain a spatial, temporal, or functional
structure~\cite{DeWolf2005}.

From these observations we develop our definition of emergent
self-organization. We reject the argument that the ability of $\TheSwarm$ to
accomplish a complex task is evidence of emergent self-organization. Consider
``programmed'' emergence in which robot behaviors ``fit'' together and directly
``sum'' to the desired outcome---clearly the collective behavior is easily
predictable from component study, and therefore not emergent.  Instead, we
define emergent self-organization as the \emph{second order} effects of the
robot control algorithm; that is, how a complex sequence of robot-environment
and robot-robot interactions can give rise to superlinear behaviors (e.g.,
performance gains)~\cite{Hunt2020}. As an example, consider a control algorithm
for cooperatively building 3D structures~\cite{Petersen2011,Nave2020}: the
swarm's ability to collectively build such structures does not in itself
indicate self-organization (a single robot can also do it). Nor does it
unilaterally indicate emergence: if a single robot takes $\TotalTime$ seconds to
build a structure, and $\TheSwarmSize$ robots take $\ge\TotalTime/\TheSwarmSize$
seconds to build the same structure, that is predictable from component study,
since increased interference might cause a slowdown.  However, if
$\TheSwarmSize$ robots can build the same structure in \emph{less} than
$\TotalTime/\TheSwarmSize$ seconds, then that \emph{is} indicative of emergent
self-organization (superlinear performance gain).

\subsection{Axes Intuitions}

Our two metric axes (see~\cref{tab:sr-taxonomy}) are based on characteristics of
self-organizing systems, in which the systems are generally far from equilibrium
and show an increase in order~\cite{DeWolf2005}.  We can infer
from~\cite{Cotsaftis2009,DeWolf2005} that SR systems with a high degree of local
interactivity should exhibit higher levels of emergent self-organization than
systems with a low degree of
interactivity~\cite{Szabo2014,Galstyan2005,Winfield2005a}.
From~\cref{fig:emergence-theory}, we know that emergent behavior can be
quantitatively measured as a super-linear increase in swarm performance as
$\TheSwarmSize$ increases linearly. We argue that sub-linear increases in
inter-robot interference across linearly increasing $\TheSwarmSize$ are
\emph{also} indicative of emergent behavior, since there may be a correlation
between reduced interference and increased performance with self-organization.

We measure \emph{spatial} self-organization, defined as collective optimization
of inter-robot interference and movement patterns, by measuring the
sub-linearity of inter-robot interference as $\TheSwarmSize$ increases. If the
proportional increase in the amount of performance \emph{lost} due to
inter-robot interference between swarms of size $\TheSwarmSize_{1}$ and
$\TheSwarmSize_{2}$, with $\TheSwarmSize_{1} < \TheSwarmSize_{2}$, is
sub-linear, then self-organization has occurred. That is, $\TheSwarm$ has learned
to leverage inter-robot interactions to become \emph{more} spatially efficient
with increasing $\TheSwarmSize$.  In the absence of self-organizing, random
robot motions would produce approximately linear increases in inter-robot
interference as the size of $\TheSwarmSize$ increases, and be in equilibrium
regarding interference levels. In order for spatial self-organization to
reliably arise, $\TheSwarm$ must be operating in a bounded space; without this
restriction, a swarm will not necessarily be motivated to manage space
efficiently. This restriction manifests itself in applications such as
warehouses~\cite{Hsieh2013}.

We measure \emph{task-based} self-organization, defined as collective
optimization of task allocation, by measuring the super-linearity of the
marginal performance gain as $\TheSwarmSize$ increases~\cite{Rosenfeld2006}. If
a super-linear increase in performance \emph{gained} due to optimization of
higher level swarm functions is observed between two swarms of size
$\TheSwarmSize_{1}$ and $\TheSwarmSize_{2}$, then emergent self-organization has
occurred. That is, the swarm has learned to leverage inter-robot interactions
and relationships between tasks to become \emph{more} efficient with larger
$\TheSwarmSize$.  In applications without a bounded operating arena, task-based
self-organization is more likely to emerge than spatial self-organization, as it
does not directly depend on environmental factors.

\subsection{Metric Derivations}
Formalizing our intuition, we define our task emergent self-organization measure
$\SpatialEmergenceMetric{\TheSwarmSize_{1}}{\TheSwarmSize_{2}}$ as follows.  We
calculate (post-hoc) the number of robots experiencing inter-robot interference,
instead of doing useful work, within $\TheSwarm$ on each timestep $t$, denoted
as $\TLost{\TheSwarmSize}$. This can then be used to compute the per-timestep
fraction of overall performance loss $\PerfLost{\TheSwarmSize}$ as follows:
\begin{equation}\label{eqn:fl-plost}
  \PerfLost{\TheSwarmSize} \mbox{$=$}
  \begin{cases}
    \PerfCurve{1}\TLost{1} & \text{if $\TheSwarmSize$\mbox{$=$}1}
    \\
    \PerfCurve{\TheSwarmSize}{\TLost{\TheSwarmSize}} \\
    - \TheSwarmSize\PerfLost{1}& \text{if $\TheSwarmSize$\mbox{$>$}1}
    \\
  \end{cases}
\end{equation}
\cref{eqn:fl-plost} quantifies a swarm's ability to detect and eliminate
non-cooperative situations between agents (e.g., frequent inter-robot
interference) as $\TheSwarmSize$ increases~\cite{George2005}. For
$\TheSwarmSize > 1$ we subtract the interference that would have occurred in a
swarm of $\TheSwarmSize$ independent robots which never interfered with each
other. If we compute~\cref{eqn:fl-plost} for two swarms of size
$\TheSwarmSize_{1} < \TheSwarmSize_{2}$, sub-linear increases in the computed
value indicate that a swarm of $\TheSwarmSize_{1}$ robots is capable of emergent
self-organization. We can now define our spatial emergent self-organization
measure $\SpatialEmergenceMetric{\TheSwarmSize_{1}}{\TheSwarmSize_{2}}$, where
positive values indicate self-organization, as:
\begin{equation}\label{eqn:pm-emergence-spatial}
  \SpatialEmergenceMetric{\TheSwarmSize_{1}}{\TheSwarmSize_{2}} = \sum_{t\in{\TotalTime}}\frac{\TheSwarmSize_{2}}{\TheSwarmSize_{1}}\PerfLost{\TheSwarmSize_{1}} - \PerfLost{\TheSwarmSize_{2}}
\end{equation}

We  define our task emergent self-organization measure
$\TaskEmergenceMetric{\TheSwarmSize_{1}}{\TheSwarmSize_{2}}$ as:
\begin{equation}\label{eqn:pm-emergence-task}
  \TaskEmergenceMetric{\TheSwarmSize_{1}}{\TheSwarmSize_{2}} = \sum_{t\in{\TotalTime}}\PerfCurve{\TheSwarmSize_{2}} -\frac{\TheSwarmSize_{2}}{\TheSwarmSize_{1}}\PerfCurve{\TheSwarmSize_{1}}
\end{equation}
\cref{eqn:pm-emergence-task} quantifies the \emph{marginal performance
  gain}~\cite{Rosenfeld2006} of the swarm as superlinear increases in
self-organization.  Positive values indicate self-organization.

%
\section{Swarm Scalability}\label{sec:meas-scalability}
\subsection{Background}\label{ssec:meas-scalability-bg}
In recent years, swarm engineering has produced many design tools and methods for
rigorous algorithm analysis and derivation of analytical, rather than weakly
inductive proofs of
correctness~\cite{Winfield2008,Matthey2009,Correll2008,Lopes2016a,Hamann2013,Moarref2018}. Despite
this, there has not been a corresponding increase in the number of robots used,
which is important for tackling problems of real-world scale.  Investigation of
simple behaviors such as pattern formation, localization, or collective motion
is generally evaluated at small scales, with 20--40 robots
\cite{Liu2009,Francesca2014,Winfield2008},
and more complex behaviors such as foraging and task allocation with 20--30
robots~\cite{Ferrante2015,Pini2011a,Correll2008}. Notable exceptions to this
trend include~\cite{Lopes2016a,Hecker2015,Hamann2008,Harwell2019a}, which used
600, 768, 375, and 16,384 robots respectively.



While these swarm sizes may technically meet the criteria to be a considered a
swarm (10--20 robots~\cite{Sahin2005}), from a swarm engineering perspective
they may not provide enough confidence in algorithm correctness, since studies at
small scales may suffer from unmanifested theoretical and empirical scalability
issues. Furthermore, reliably analyzing emergent and self-organizing behaviors
during the design process is difficult at small scales since they only reliably
arise at larger scales, where ``larger'' is
problem-dependent~\cite{Li2016}.

Previous work calculated swarm scalability as
$S(\TheSwarmSize)=\PerfMeasureHecker{\TheSwarmSize}/\TheSwarmSize$ (i.e.,
per-robot efficiency)~\cite{Hecker2015}. While this measure provides insight
into scalability, it is not predictive (e.g.~given
$\PerfMeasureHecker{\TheSwarmSize}$, we cannot plausibly estimate
$\PerfMeasureHecker{{2}\TheSwarmSize}$ without retroactively charting
$S(\TheSwarmSize)$ across a range of values for $\TheSwarmSize$).

\subsection{Axis Intuition}
Our metric axis (see~\cref{tab:sr-taxonomy}) is based on the intuition that
$\TheSwarmSize$ robots in $\TheSwarm$ are analogous to nodes in a supercomputing
system: if a job takes $T$ seconds with $\TheSwarmSize$ robots, ideally it will
take $T/2$ seconds with $2\TheSwarmSize$ robots in a perfectly parallelizable
system. We adapt the Karp-Flatt metric~\cite{Karp1990} which measures the level
of parallelization of a program and the plausibility of speedups if more
computational resources are added, to swarm engineering, where it exposes the
portion of observed performance rooted in inter-robot cooperation. Using such a
metric, higher values will indicate cooperative swarms and plausible predictions
of efficiency increases for larger $\TheSwarmSize$.

\subsection{Metric Derivation}
Formalizing our intuition, we define our scalability measure
$\ScalabilityMetric{\TheSwarmSize_{1},\TheSwarmSize_{2}}$, where
$\TheSwarmSize_{1} < \TheSwarmSize_{2}$, using the Karp-Flatt metric:
\begin{equation}
  \label{eqn:pm-scalability}
  \ScalabilityMetric{\TheSwarmSize_{1},\TheSwarmSize_{2}} = \sum_{t\in{\TotalTime}}1 -
  \Bigg[\frac{\frac{\PerfCurve{\TheSwarmSize_{2}}}{\PerfCurve{\TheSwarmSize_{1}}} -
    \frac{1}{\TheSwarmSize_{2}/\TheSwarmSize_{1}}}{1- \frac{1}{\TheSwarmSize_{2} / \TheSwarmSize_{1}}}\Bigg]
\end{equation}
In \cref{eqn:pm-scalability}, the term in $[~]$ is the serial fraction
$\mathbf{e}$ from the original Karp---Flatt paper. Our formulation defines
``speedup'' as the performance gain with $\TheSwarmSize_{2}$ robots over the
performance with $\TheSwarmSize_{1}$ robots, rather than over the performance
with 1 robot, per the original paper. This ``marginal performance gain''
provides comparable results between $\kappa$ which require a critical
$\TheSwarmSize$ to perform well (i.e., sub-linear increases in performance
across small $\TheSwarmSize$), and $\kappa$ which have more linear increases in
performance across small $\TheSwarmSize$. Negative values for
$\ScalabilityMetric{\TheSwarmSize_{1},\TheSwarmSize_{2}}$ are possible, due to
the fact that the original Karp-Flatt metric assumes that
$\PerfCurve{\TheSwarmSize}$ is a monotonically increasing function (i.e., adding
more processors always gives some marginal gain), and this assumption is not
valid in general for SR systems when adding more robots.

%
%
\section{Swarm Flexibility}\label{sec:meas-flexibility}
\subsection{Background}\label{ssec:meas-flexibility-bg}
\emph{Flexibility} among agents within a swarm in response to environmental
changes is an important aspect of their collective utility: they should be able
to react to factors in the environment, such as the quality or safety of
locations within it, while also not changing their behavior in response to every
fluctuation~\cite{Hunt2020}.
Some works on flexibility compare the raw performance of swarm control
algorithms under dynamic environmental conditions~\cite{Just2017,Lerman2006} or
conditions from those they were developed
in~\cite{Duarte2016,Ferrante2013a,Hecker2015}. However, the resultant claims of
flexibility due to performance similarity across scenarios is strictly
qualitative, due to a lack of mathematical quantification of ``difference'' in
scenario conditions as a factor in performance comparisons.
\begin{figure}[ht]
  {\includegraphics[width=0.46\linewidth]{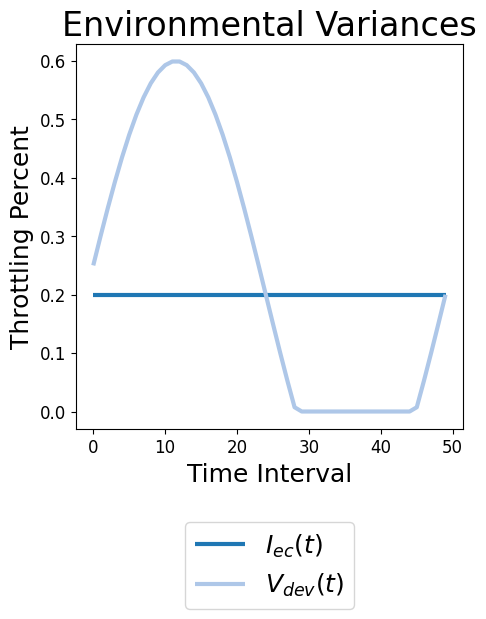}}
  {\includegraphics[width=0.53\linewidth]{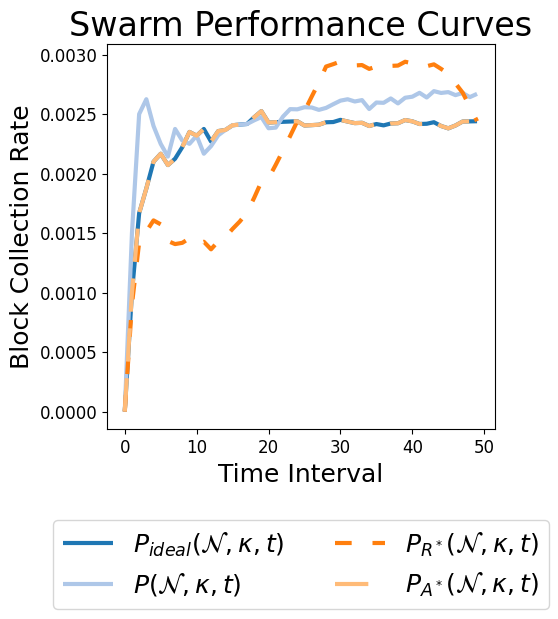}}%
  \caption{\label{fig:flexibility-ex}\footnotesize{Environmental conditions
      modeled as a sinusoidal deviation ($\NonIdealEnvDeviation$) from ideal
      conditions ($\IdealEnvConditions=0$) (left), and the resulting swarm
      behavior curves (right). Such conditions can arise, for instance, in
      outdoor scenarios with periodic winds hampering robot
      motion. $\PerfCurveSub{\TheSwarmSize}{ideal}$ is the swarm performance
      under $\IdealEnvConditions$, and $\PerfCurve{\TheSwarmSize}$ is the swarm
      performance under $\NonIdealEnvDeviation$. $\OptimalReactivityCurve$ is
      the ideal reactivity curve, and $\OptimalAdaptabilityCurve$ is the ideal
      adaptability curve.}}
\end{figure}

\subsection{Axes Intuitions}
A swarm $\TheSwarm$ is considered flexible if it can self-organize to cope with
a variety of non-ideal environmental conditions (defined by
$\NonIdealEnvDeviation$) which change over time and are different than the
conditions it was developed in. These changes can include the cost of performing
a particular action (e.g., picking up/dropping an object in foraging), maximum
robot speed (e.g., wheeled robots buffeted in variable winds in outdoor
environments) or the performance of some tasks more slowly than others (e.g.,
carrying objects of differing weights).

We measure swarm flexibility along two opposing axes
(see~\cref{tab:sr-taxonomy}) in order to encompass a wide range of possible
design constraints. First, its \emph{reactivity}: how quickly and proportionally
$\PerfCurve{\TheSwarmSize}$ changes according to changes in
$\NonIdealEnvDeviation$, as shown in~\cref{fig:flexibility-ex}. Second, its
\emph{adaptability}: $\PerfCurve{\TheSwarmSize}$ does not change, regardless of
$\NonIdealEnvDeviation$ (i.e., consistent performance). That is, if non-ideal
environmental conditions change over time, performance does \emph{not} change in
ideal environmental conditions, as shown in~\cref{fig:flexibility-ex}. Adaptive
swarms minimize performance losses under adverse conditions
($\NonIdealEnvDeviation > 0$), and resist performance increases under beneficial
deviations from $\IdealEnvConditions$, (i.e., $\NonIdealEnvDeviation < 0$).

\subsection{Metric Derivations}
Using the results of~\cite{Jekel2018}, we derive temporally-aware measures
quantifying the difference between expected and observed performance in
arbitrary environmental conditions, and use it to calculate swarm reactivity and
adaptability.

To derive the performance curve of a swarm with optimal reactivity,
$\OptimalReactivity$, we note that such a swarm will react \emph{instantaneously} and
\emph{proportionally} whenever $\NonIdealEnvDeviation$ is non-zero. Define $c_t$ as a
per-timestep constant representing the signed proportional difference between ideal
and non-ideal conditions:
\begin{equation}\label{eqn:reactivity-ct}
  c_t = \frac{\NonIdealEnvDeviation + \IdealEnvConditions}{\IdealEnvConditions}
\end{equation}
Then, $\OptimalReactivityCurve = c_t\PerfCurveSub{\TheSwarmSize}{ideal}$ for all
$t\in{T}$, and we can now define our reactivity measure $\ReactivityMetric$:
\begin{equation}\label{eqn:reactivity}
  \ReactivityMetric = DTW(\OptimalReactivityCurve,\PerfCurve{\TheSwarmSize})
\end{equation}
where $DTW(X,Y)$ is a \textit{Dynamic Time Warp} similarity measure between two
discrete curves $X$ and $Y$, based on the conclusions in~\cite{Jekel2018}. We note
that  DTW correctly matches temporally shifted sequences of applied variance and
observed performance (achieving $\OptimalReactivity$ is not possible for realistic
swarms), and exhibits robustness to signal noise.

It is clear from our intuition above that in a swarm with optimal adaptability,
$\OptimalAdaptability$
$\OptimalAdaptabilityCurve=\PerfCurveSub{\TheSwarmSize}{ideal}$ for all $t$. We
can now define our adaptability measure $\AdaptabilityMetric$:
\begin{equation}\label{eqn:adaptability}
  \AdaptabilityMetric = DTW(\OptimalAdaptabilityCurve,\PerfCurve{\TheSwarmSize})
\end{equation}
where $DTW(X,Y)$ is a \textit{Dynamic Time Warp} similarity measure, for the same
reasons described above for $\ReactivityMetric$.

%
\section{Swarm Robustness}\label{sec:meas-robustness}
\subsection{Background}\label{ssec:meas-robustness-bg}
Bjkernes et al.~\cite{Bjerknes2013} show that contrary to the common assumption
in SR literature, larger swarms can be less robust than smaller swarms, and
argue that fundamental changes to SR system design, such as the inclusion of an
artificial immune system, are required to detect and self-heal from faults. A
generic framework for fault detection in robot swarms based on immunological
response and peer voting was presented in~\cite{Tarapore2017}, and showed a high
level of accuracy in identifying a faulty robot in real-robot swarms, even if
swarm behavior collectively changed over time. Other works in formation
control~\cite{GuerreroBonilla2020b} and decision making~\cite{Usevitch2019}
treat robustness from the perspective of the ability of $\TheSwarm$ to meet its
goals in the presence of malicious robots, and provide provable collective
behavior guarantees, which are by their nature guarantees of emergent behavior.
%
Availability analysis, i.e., determining what fraction of a swarm will be
available on average in statistical equilibrium to work on tasks, given robot
malfunction and repair rates, has investigated both independent and coupled
robot malfunction rates~\cite{Zhang2008}.

Swarm robustness in the presence of single robot sensor malfunction was
demonstrated in~\cite{Ramachandran2020}, allowing failing robots to share data
with functioning robots. Swarm robustness to Gaussian noise
was measured and analyzed by~\cite{Turgut2008,Dallalibera2011} on collective
swarm motion, and on obstacle detection~\cite{Claudi2014}. Gaussian noise was
applied to robot sensor readings in~\cite{Castello2016}, and their effects
studied for different variants of a foraging algorithm. The Fokker-Planck
equation was derived from statistical physics for a given Langevin equation
governing robot motion, including noise models, and applied to the analysis of
simple phototaxis (i.e., motion towards light)
behaviors~\cite{Hamann2008}. Other
works 
inject a small level of uniform or Gaussian noise (0.5--10\%) to help narrow the
simulation-reality gap, without performing statistical
analysis~\cite{Hecker2015,Pini2011a}. Many theoretic works do not incorporate
noise into their models~\cite{Lerman2006,Campo2007}; exceptions
include~\cite{Turgut2008}. Similarly, many application oriented works do not
systematically evaluate algorithms in the presence of noise;
exceptions include~\cite{Castello2016,Dallalibera2011}.

\subsection{Axes Intuitions}
%
We motivate our robustness axes by considering real world examples such as
underwater, mining, and warehouses~\cite{Duarte2016,Hsieh2013}.  In such
environments, swarms may have to deal with one or more of: (1) high levels of
sensor and actuator noise, (2) adversarial task re-allocation in which robots
can leave a sub-swarm working on a certain task at any time, and (3) frequent
robot mechanical failures. We therefore define a swarm's \emph{robustness} along
two axes.

First, \emph{sensor and actuator noise robustness}, or its ability to
collectively adapt to temporally varying levels of sensor and actuator noise.
Swarm sensor and actuator noise robustness should measure how close
$\PerfCurve{\TheSwarmSize}$ of $\TheSwarm$ remains to
$\PerfCurveSub{\TheSwarmSize}{ideal}$ when $\NonIdealEnvDeviation$ affects the
\emph{internal} state of $\TheSwarm$ via sensor and actuator noise, where
$\PerfCurveSub{\TheSwarmSize}{ideal}$ is the performance of $\TheSwarm$ with
idealized, noise-free sensors.

Second, \emph{population dynamics robustness}, or its ability to collectively
adapt to temporal changes in the swarm size $\TheSwarmSize$. We emphasize the
\emph{collective} adaptation, as our robustness definition is an emergent
property.  Swarm population dynamics robustness of $\TheSwarm$ should measure
the ability to maintain the same level of performance as population size
changes.  As robots enter and leave $\TaskedSwarm$ over time (permanently or
temporarily), they will have an inaccurate world model. Different $\kappa$ will
cope with these robots in different ways and will ultimately affect
$\PerfCurve{\TaskedSwarmSize{t}}$ differently.  We want to correlate
$\TaskedSwarmSize{t}$ with changes in $\PerfCurve{\TaskedSwarmSize{t}}$. Building
on~\cite{Zhang2008}, we derive expressions for the average number of robots in
the system and how long a robot stays in the system, and use this to define our
population dynamics robustness measure.

\subsection{Metric Derivations}
It is clear from our intuition above that in a swarm with optimal sensor and
actuator noise robustness $\OptimalSARobustness$,
$\OptimalSARobustnessCurve=\PerfCurveSub{\TheSwarmSize}{ideal}$ for all $t$. We
can now define our sensor and actuator noise robustness measure
$\SARobustnessMetric$:
\begin{equation}\label{eqn:sa-robustness}
  \SARobustnessMetric = DTW(\OptimalSARobustnessCurve,\PerfCurve{\TheSwarmSize})
\end{equation}
where $DTW(X,Y)$ is a \textit{Dynamic Time Warp} similarity measure, as
described in~\cref{sec:meas-flexibility} for $\ReactivityMetric$ and
$\AdaptabilityMetric$. We compare performance \emph{curves}, rather than
cumulative performance values, in order to account for swarm emergent (partial)
mitigation of the noise via changing behaviors.

To compute the variability of $\TaskedSwarmSize{t}$, we first derive the average
amount of time a robot spends in $\TaskedSwarm$, $\TimeInTaskedSwarm$.

In queueing theory, queues are described by a tuple $A/S/c$~\cite{Seda2017}. $A$
is the arrival process, which is typically Markovian, $S$ is the service time
distribution, which is also typically Markovian, $c$ is the number of service
channels available to service items as they leave the queue. We study the
$M/M/1$ queue, in which items arrive at rate $\sim{Poisson(\lambda)}$, remain in
the queue for a period of time, and are serviced at rate
$\sim{Exp(\mu)}$~\cite{Seda2017}. We define three queues to model the possible
sources of a temporally varying $\TaskedSwarmSize{t}$ resulting from robots
leaving $\TaskedSwarm$, extending~\cite{Zhang2008}, as follows:
\begin{enumerate}
\item {\emph{Birth queue}, $\BirthQueue$, is an M/M/1 queue defined by
    $(\lambda_{b}=0,\mu_{b})$, which models the periodic addition of robots to
    $\TheSwarm$ at rate $\mu_{b}$ that have been released from other tasks to
    join $\TaskedSwarm$ to work on task $\AssignedTask$. $\BirthQueue$ begins
    with $(\TheSwarmSize-\TaskedSwarmSize{0})$ robots in it and has a finite
    population; that is, every robot not part of $\TaskedSwarm$ at $t=0$ is a
    member of this queue.}
\item {\emph{Death queue}, $\DeathQueue$, is an M/M/0 queue defined by
    $(\lambda_{d},\mu_{d}=0)$ modeling the permanent removal of robots from
    $\TaskedSwarm$ as they critically malfunction or are permanently reallocated
    to other tasks and are removed from the swarm at rate $\lambda_{d}$. We
    treat the two cases identically. There are 0 servers in this queue, and
    hence $\mu_{d}=0$. $\DeathQueue$ begins with 0 robots in it.}
\item {\emph{Repair/reallocation queue}, $\RepairQueue$, is a birth-death M/M/1
    queue defined by $(\lambda_{bd},\mu_{bd})$. $\RepairQueue$ models the
    \emph{temporary} removal of robots from $\TaskedSwarm$ as they (1)
    non-critically malfunction, or (2) are temporarily reallocated to other
    tasks at rate $\lambda_{bd}$ (we treat the two cases identically), and
    return to $\TaskedSwarm$ to work on $\AssignedTask$ at rate $\mu_{bd}$.
    $\RepairQueue$ begins with 0 robots in it.}
\end{enumerate}
Assuming that (1) the service/arrival rates are constant, (2)
$\BirthQueue,\DeathQueue,\RepairQueue$ are all independent, and (3) a robot can
be at most in one queue at a time, we leverage the additive property of Poisson
random variables and combine $\BirthQueue,\DeathQueue,\RepairQueue$ into a
single M/M/1/$\TaskedSwarmSize{t}$/$\TheSwarmSize$ queue, $\UntaskedSwarmQueue$,
which models the number of robots \emph{not} in $\TaskedSwarm$ at $t$.

To have a stable system, the \emph{utilization} $\rho_{\bar{\TaskedSwarm}}$ must be
$<1$:
\begin{equation}\label{eqn:robustness-rho}
  \rho_{\bar{\TaskedSwarm}} = \frac{\UntaskedSwarmQueueLambda}{\UntaskedSwarmQueueMu}
\end{equation}
Depending on the relative arrival and service rates of
$\BirthQueue,\DeathQueue,\RepairQueue$, stability can be achieved in many
different ways. Similarly, the variance of the size of $\UntaskedSwarmQueue$ can
also vary greatly as robots stay in the swarm for different amounts of time on
average. Using Little's law~\cite{Seda2017} to calculate the average amount of
time each robot spends in $\UntaskedSwarmQueue$ we have:
\begin{equation}\label{eqn:robustness-ss-queue-length}
  L_{\UntaskedSwarm} = \frac{{\rho_{\UntaskedSwarm}}^2}{1 - \rho_{\UntaskedSwarm}}
\end{equation}
and the total amount of time $\TimeInUntaskedSwarm$ that a robot spends not in
$\TaskedSwarm$:
\begin{equation}
  \TimeInUntaskedSwarm = \frac{1}{\UntaskedSwarmQueueMu - (\UntaskedSwarmQueueLambda)} + \frac{1}{\UntaskedSwarmQueueMu}
\end{equation}
We can now compute the amount of time a robot is \emph{not} in
$\UntaskedSwarmQueue$ (and therefore in $\TaskedSwarm$), by taking
$\TimeInTaskedSwarm = \TotalTime - \TimeInUntaskedSwarm$. Higher values of
$\TimeInTaskedSwarm$ indicate more stable swarms, and lower values indicate more
volatile swarms in which more ``experienced'' robots will have to work with the
sub-optimal actions of less experienced robots which are newly added, have been
repaired, or have finished a different task.  We now define our population
dynamics robustness measure $\PDRobustnessMetric$ by correlating
$\TimeInTaskedSwarm$ with $\PerfCurve{\TaskedSwarmSize{t}}$:
\begin{equation}\label{eqn:size-robustness}
  \PDRobustnessMetric = \sum_{t\in{\TotalTime}}\PerfCurve{\TaskedSwarmSize{t}} - \frac{\TimeInTaskedSwarm}{\TimeInTaskedSwarmIdeal}\PerfCurveSub{\TaskedSwarmSize{t}}{ideal}
\end{equation}
\cref{eqn:size-robustness} measures the difference between
$\PerfCurveSub{\TaskedSwarmSize{t}}{ideal}$ and $\PerfCurve{\TaskedSwarmSize{t}}$,
weighted by swarm population variance,
$\TimeInTaskedSwarm/\TimeInTaskedSwarmIdeal$.

\section{Application Scenarios Overview}\label{sec:sc-overview}

We examine two scenarios: indoor warehouse object transport (\cref{sec:sc1}),
and outdoor search and rescue (\cref{sec:sc2}), framed as swarm control
algorithm selection problems. The scenarios span different problem domains in
which SR systems have been successfully applied, and thus serve as excellent
testbeds for determining the utility of our approach as an assistive tool in the
development of SR systems. Both scenarios are ``obstacle-free'', in that there
are no predefined obstacles. However, robots act as moving obstacles to each
other; this maps naturally to many SR applications in which there are no humans
in the operating area due to safety concerns.

We frame our two scenarios as \emph{foraging tasks}, in which robots gather
objects from a finite operating arena and bring them to a central location.  We
include intuitions about how each controller will perform in each scenario based
solely on their technical descriptions (\cref{sec:sc1,sec:sc2}, respectively),
so that we can determine to what extent our intuitions are supported by the
numerical results from our metrics.


\subsection{Motivation via Modeling Scenario Characteristics}\label{ssec:sc-overview-modeling}
We motivate the need for insightful metrics by considering the design
constraints shown below, each of which may be present or absent in specific
scenarios such as the two we study in this work. Without precise numerical
quantification of the effect of constraints on swarm behavior, then any
evaluation of control algorithms will be strictly qualitative, and based solely
on iterative refinement and engineer intuition.  We show that our metrics
encompass all the constraints shown below, and therefore solve the problem of
modeling real-world scenario characteristics in a comprehensive way.
\begin{enumerate}
\item

  \emph{Environmental Disturbances}. In many cases, it may not be feasible to
  develop and test SR systems in the exact environmental conditions in which
  they will operate (e.g., space, military, disaster cleanup applications)
  due to cost or locality constraints~\cite{Rouff2007a}. Therefore, flexibility in
  reacting and adapting to unknown conditions is critical.

\item

  \emph{Noisy environments}. In many environments, sensor and actuator readings
  may not be accurate beyond the usual level of noise. Levels of noise can vary
  (1) depending on the robot's current location in the operating arena, and (2)
  the current time (e.g., temporally varying sources). In this work, we consider
  a noise model in which the exact noise level is not known \emph{a priori}, but
  its distribution is known or can be reasonably inferred.  We further assume
  that the effects of sensor and actuator noise are spatially homogeneous and
  affect all sensors and actuators equally.

\item

  \emph{Unreliable robots}. All robots are fallible, due to the nature of
  robotic hardware. It is only a question of how often they fail and if the
  failure is permanent or repairable. Some prior work has considered temporary
  failures~\cite{Zhang2008,Ramachandran2020}; we expand the scope by considering
  the effect of temporary \emph{and} permanent robot failures.

\item

  \emph{Semi-adversarial task reallocation}. Given the inherent flexibility to
  adapt to perform different tasks efficiently, it is expected that not all
  robots within a swarm will be working on the same task at the same time, as
  different system operators dynamically reallocate robots to tasks in
  real-time. Previous work in swarm robotics in general, and foraging in
  particular, assumed that system operators will have exclusive use of the
  entire swarm, whatever its size is, for the task at hand (i.e.,
  $\TheSwarmSize=\TaskedSwarmSize{t}$ for all $t$)
  (\cite{Harwell2018,Harwell2019a,Harwell2020a,Pini2011a,Hecker2015}). For
  simplicity of analysis, we assume that all robot reallocations are independent
  and can happen at any time.
\item

  \emph{Operating arena size}. If the size of the arena in which $\TheSwarm$
  will operate is known beforehand, \emph{variable swarm density} is an
  appropriate model, in which swarms of increasing size $\TheSwarmSize$ are
  placed in the same arena. If the size of the arena is not known beforehand,
  \emph{constant swarm density} is more appropriate, in which the size of the
  operating arena scales linearly with increasing $\TheSwarmSize$. Our
  definition of swarm density loosely falls under the Eulerian continuum for
  swarm dynamics described in~\cite{Gazi2004}.
  The choice of variable or constant swarm density is a critical factor in the
  design process. With variable density, robots in swarms with higher densities
  will experience \emph{more} inter-robot interactions than robots in smaller
  swarms, thus skewing performance results.  With constant density, as swarm
  sizes increase, the arena size also increases proportionally, greatly reducing
  skewing artifacts. An appropriate model needs to be chosen in either case, as
  swarm density, not size, is the critical factor in determining swarm emergent
  self-organization~\cite{Sugawara1997,Hamann2013}.
\item \emph{Object Distribution}. The distribution of the objects to be gathered
  in the operating arena can be modeled in various ways. The most common
  distribution is \emph{random} (\cref{fig:rn-foraging}), in which objects are
  scattered uniformly in a square or circular
  environment~\cite{Campo2007,Sugawara1997,Castello2016,Hecker2015}.  Other
  studied distributions include \emph{power law} (\cref{fig:pl-foraging}), in
  which objects are clustered in groups of various sizes~\cite{Hecker2015}. When
  the object distribution is not known or cannot be inferred \emph{a priori},
  \emph{random} or \emph{power law} are appropriate, following empirical
  observations~\cite{Hecker2015}.  The \emph{single source} distribution
  (\cref{fig:ss-foraging}) has also been extensively
  studied~\cite{Harwell2018,Harwell2019a,Ferrante2015,Harwell2020a,Pini2011a}.

\begin{figure}[ht]
  \subfloat[\label{fig:rn-foraging} Example \emph{random} object distribution
  (scenario 1 medium warehouse,~\cref{sec:sc1}).]
  {\includegraphics[width=.48\linewidth]{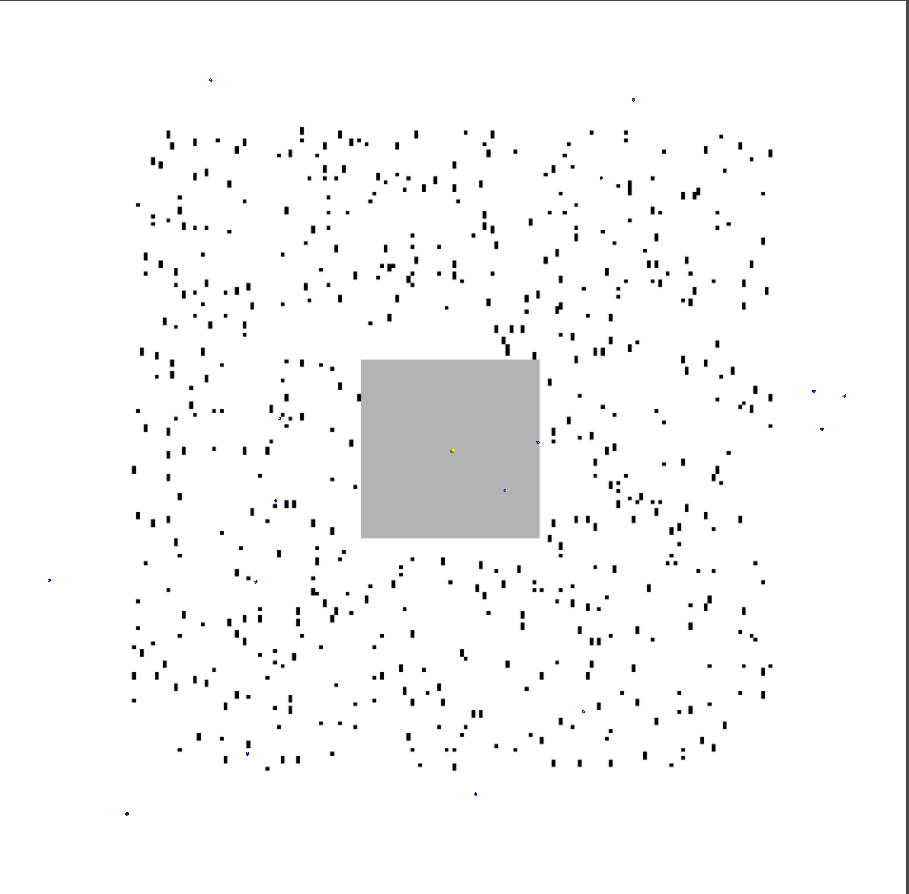}}%
  \hfill
  \subfloat[\label{fig:pl-foraging}Example \emph{power law } object distribution
  (scenario 2,~\cref{sec:sc2}).]
  {\includegraphics[width=.48\linewidth]{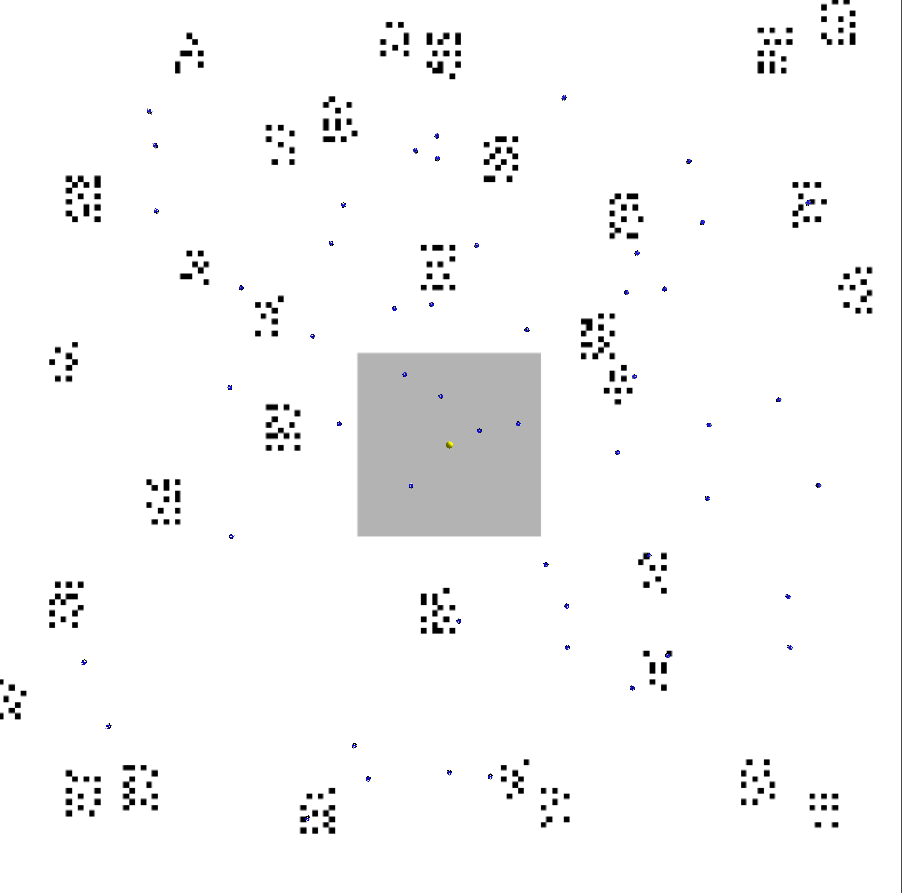}}\label{fig:foraging-dists}
  \newline
  \subfloat[\label{fig:ss-foraging}Example \emph{single source} object distribution
    (scenario 1 small warehouse,~\cref{sec:sc1}).]
  {\includegraphics[width=0.99\linewidth,height=0.15\textwidth]{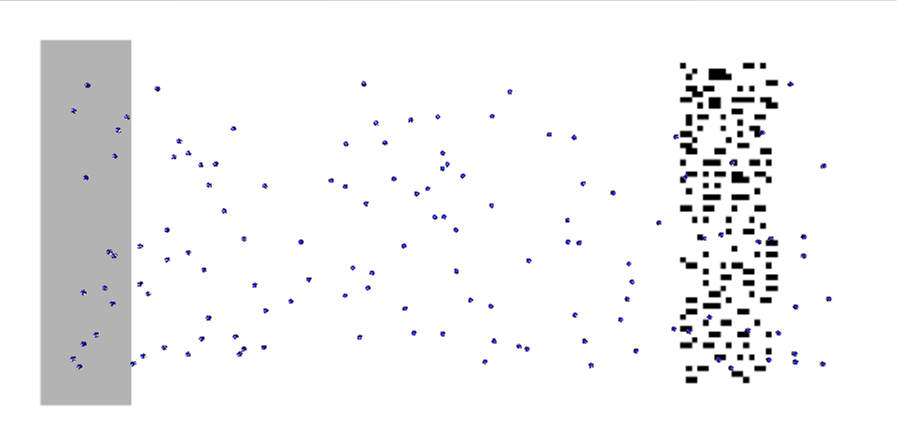}}
  \caption{\footnotesize{Examples of
      foraging distributions. The nest is dark gray, objects are black, robots dark blue, and the
      yellow blobs are lights above the nest which robots use for
      phototaxis.}}
\end{figure}
\item \emph{Minimum collective performance}. In SR systems with task
  reallocation and robotic failures, a common design constraint is a minimum
  performance that $\TaskedSwarm\in\TheSwarm$ must meet in statistical
  equilibrium, as $\TaskedSwarmSize{t}$ fluctuates. We use the notion of
  \emph{swarm availability}, and define $p_v$ as the probability that
  $\TaskedSwarmSizeMin$ robots from a swarm of size $\TheSwarmSize$ are
  allocated to task $\AssignedTask$ at time $t$ (for a derivation,
  see~\cite{Zhang2008}):
\begin{equation}\label{eqn:swarm-availability}
  \begin{aligned}
    p_v =& \pi_{\TheSwarmSize}\big(1+\sum_{k=\TaskedSwarmSizeMin}^{\TheSwarmSize-1}\prod_{i=k+1}^{\TheSwarmSize}\frac{1}{\rho}\big)
  \end{aligned}
\end{equation}
Using~\cref{eqn:swarm-availability} we can now solve for either:
\begin{itemize}
\item {$\rho$, obtaining a \emph{ratio} that the rates of robots entering/leaving
    $\TheSwarm$ must satisfy if $(p_{v},\TaskedSwarmSizeMin)$ are specified. }
\item {A range of achievable $p_v$, given the rates of robots entering/leaving
    $\TheSwarm$, and a range of $\TaskedSwarmSizeMin$ values
    (see~\cref{fig:sc1-availability-small} for examples of this analysis).}
\end{itemize}
\end{enumerate}
\subsection{Candidate Algorithms ($\kappa$)}\label{ssec:sc-candidate-algs}
We have selected four state-of-the-art foraging swarm control algorithms, which
in our notation we indicate with $\kappa$.  Videos for each $\kappa$ with
different spatial distributions of the swarm are included in the multimedia
accompaniment to this paper, to provide intuitive grounding of the behavior of
each
algorithm\footnote{\url{https://www-users.cs.umn.edu/~harwe006/showcase/tro-2021}}.
We selected these algorithms because they use common paradigms for foraging
algorithms, from simple reactive random walks to more complex methods with
theoretical bases which do task allocation. Comparative analysis across such a
wide spectrum of complexity using our metrics will constitute a significant
``stress test.'' If they provide predictive insights, we will have strong
empirical evidence that they capture underlying elements of swarm behavior.

\begin{itemize}
\item \textbf{Correlated Random Walk (CRW)}. Robots do a correlated random walk,
  which is a random walk in which the direction of the motion at each step
  depends on the direction of the previous step~\cite{Renshaw1981}, until they
  acquire an object, which they then transport to the nest using phototaxis,
  i.e., motion in response to light. Robots do not reallocate tasks, i.e., they
  always do the same task, and have no memory (reactive algorithm). Similar
  controllers can be found in~\cite{Lerman2001,Galstyan2005}.

\item

  \textbf{Decaying Pheromone Object (DPO)}.  Robots track objects they have seen
  using exponentially decaying pheromones~\cite{Hecker2015,Harwell2020a}, and
  determine the ``best'' object to acquire using derived information
  relevance. Robots do not perform dynamic task allocation (i.e., always perform
  the same task). Similar controllers can be found in~\cite{Talamali2020}.
  \begin{figure}[ht]
    \centering
    \includegraphics[width=0.98\linewidth]{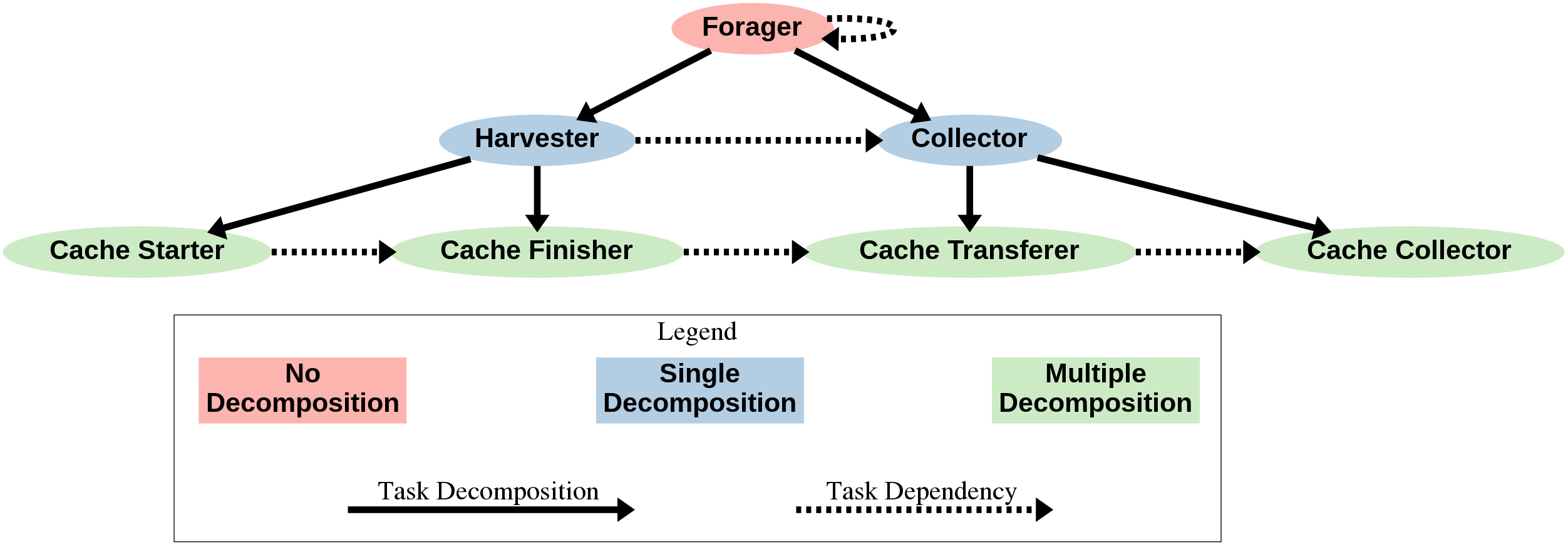}
    \caption{\label{fig:tdgraph-foraging} \footnotesize{Task decomposition graph
        for the foraging task. The set of red and blue tasks correspond to a
        scheme in which robots do the whole task or 1/2 task (single
        decomposition option).  The set of red and blue and green tasks
        correspond to a scheme in which robots do the whole task, 1/2 task, or
        1/4 task (multi-decomposition option).  By decomposing the task into
        multiple parts, there are now task dependencies between the parts, which
        the swarm has to collectively learn.  }}
  \end{figure}

\item \textbf{STOCHM}. Robots stochastically allocate tasks using the STOCH-N1
  algorithm~\cite{Harwell2020a}, a stochastic choice depending on the location
  of the most recently executed task within the graph of tasks available to each
  robot (\cref{fig:tdgraph-foraging}). From~\cref{fig:tdgraph-foraging}, STOCHM
  robots can allocate any of the red or blue tasks, which are: (1) the entire
  task, \emph{Forager}, retrieving an object and bringing it to the nest, (2)
  the first half of the task, \emph{Harvester}, retrieving an object and
  bringing it to a \emph{static} intermediate drop site (cache), or (3) the
  second half of the task, \emph{Collector}, picking up a block from a cache and
  bringing it to the nest.

\item
  
  \textbf{STOCHX}. Robots also use the STOCH-N1 algorithm on the graph
  in~\cref{fig:tdgraph-foraging}. However, they can now choose red, blue, or
  green tasks, i.e., they can exploit existing caches \emph{and} perform tasks
  related to dynamically creating new caches. The new available tasks are:
  placing a block somewhere to start a new cache, \emph{Cache Starter}, placing
  a block next to a started cache to finish creating it, \emph{Cache Finisher},
  transferring a block between two caches, making progress towards the nest,
  \emph{Cache Transferer}, and picking up a block from a cache and bringing it
  to the nest, \emph{Cache Collector}. A similar algorithm for mobile caches was
  developed in~\cite{Lu2020}.
\end{itemize}

%
\subsection{Experimental Framework}\label{sec:exp-framework}
The experiments described in this paper have been carried out in the ARGoS
simulator~\cite{Pinciroli2012}. We employ a dynamical physics model of the
robots in a three dimensional space for maximum fidelity, even if robots are
restricted to motion in the XY plane, using a model of the marXbot developed
by~\cite{Dorigo2005c}. For all the experiments we average the results of 192
experimental runs of $\TotalTime = 10,000$ seconds. We utilize
SIERRA\footnote{\url{https://github.com/swarm-robotics/sierra.git}} to
automatically generate simulation inputs, run experiments, and generate the
graphs seen in this paper.

\section{Scenario 1: Indoor Warehouse}\label{sec:sc1}
This scenario is inspired by the use of robots in Amazon warehouses, in which SR
system designers have to select a control algorithm to be deployed in a small
($\sim{500~m^{2}}$) and a medium ($\sim{2000~m^{2}}$) warehouse of an order
fulfillment company. The company has just expanded from a previous small
warehouse, where they had 50 robots divided into 10 teams of 5, with each team
assigned to assist with filling a given order.

After some preliminary analysis, the company believes that sufficient
throughput can be maintained with $\TaskedSwarmSizeMin=50$ robots in the
small warehouse, but do not know how many robots will be needed in the
medium warehouse.


\begin{table}[ht]
  \centering\scriptsize
  \begin{tabularx}{\linewidth}{@{}p{.75cm}YY@{}}
    $\kappa$ &  Intuitions \\

    \toprule
    CRW &
    \begin{minipage}[t]{\linewidth}
      \begin{itemize}
      \item {Modest spatial self-organization compared to other algorithms due
          to the efficiency of random motion in reducing interference. }
      \item {Moderate reactivity and adaptability as a memory-less algorithm.}
      \end{itemize}
    \end{minipage} \\

    \midrule
    DPO &
    \begin{minipage}[t]{\linewidth}
      \begin{itemize}
      \item {Low spatial self-organization compared to CRW swarms because of
          competition at collectively learned locations of foraging targets. }
      \item {Low reactivity and adaptability because $\TheSwarm$ is affected by
          changing environmental conditions but cannot substantially change
          behavior.}
        \end{itemize}
      \end{minipage} \\

    \midrule
    STOCHM &

    \begin{minipage}[t]{\linewidth}
    \begin{itemize}
    \item {Suffers from congestion near the locations of static caches, which
        may outweigh the self-organization possible due to multiple tasks. }
    \item {Low self-organization because algorithm complexity and spatial cache requirements make
        learning through interaction in spatially constrained environments
        difficult.}
    \item {More flexible than DPO due to the presence of
        caches to help regulate traffic, and the maintenance of the caches
        by an outside process independent of swarm behavior.}
      \item {Less scalable than CRW and DPO because it balances
          exploitation and exploration of the search space.}
    \end{itemize}
    \end{minipage}\\

    \midrule
    STOCHX &

    \begin{minipage}[t]{\linewidth}
    \begin{itemize}
    \item {High levels of self-organization, greater flexibility, and population
        dynamics robustness via dynamic cache usage. }
    \item {Algorithm complexity and spatial cache management make learning
        in spatially constrained environments difficult, so
        high task-based self-organization only expected in less constrained
        environments. }
    \item {Lower raw performance than other algorithms because $\TheSwarm$ has
        to do additional work on dynamic cache management (e.g.,
        exploration vs.~exploitation of the environment).}
    \item {Less scalable than STOCHM because the solution space is larger.}
    \end{itemize}
    \end{minipage} \\

    \bottomrule

  \end{tabularx}\caption{\footnotesize{Summary of control algorithms 
      intuitions, along with comparative analysis in the indoor
      warehouse object transport scenario.}}\label{tab:sc1-kappa-summary}
\end{table}
\subsection{Design Constraints}\hfill

This scenario, or variants of it, has been studied
by~\cite{Harwell2018,Harwell2020a,Pini2011a,Castello2016}. Given the
company's description of their scenario, we define the following design constraints:
\begin{itemize}
\item { $\PerfCurve{\TheSwarmSize}$: Measures the rate of object collection; more
    efficient $\kappa$ gather more objects per $t$ on average. }
\item { Operating arena: Fixed sizes ($500~m^{2}$, $2,000~m^{2}$) with variable swarm
    density. }
\item { Object distribution: Single source for the small warehouse, and random for
    the medium warehouse.}
\item { Environmental disturbances: Periodic events such as shift changes and
    deliveries may occur, which we model as square waves, applying throttling
    functions to the maximum robot speed while carrying an object. This models
    increased congestion or additional computational time needed to plan safe
    trajectories in the presence of unknown dynamic obstacles while transporting
    goods.

    Given that the frequency and disruption of the disturbances is unknown, we
    test throttling levels $0-0.40$, and set the frequency $f=5000$, modeling
    periods of intense disruption that appear roughly every 1.4 hours.}

\item { Sensor and actuator noise: Given the tightly controlled indoor
    environment, there are no unexpected sources of sensor or actuator noise;
    precise indoor localization methods may also be available. However, no SR
    system is noise free, so we will investigate the effect of minor levels of
    Gaussian noise $G(\mu,\sigma)$ with $\mu=0,\sigma=0-0.03$. }
\item { Task reallocations: Given that robots will work alongside humans in both
    warehouses, we assume that human employees (or a centralized warehouse AI)
    can reallocate robots at any time as workloads evolve throughout the day. We
    do not have an estimate of how often this will occur from the company's
    problem description, but given that $\TaskedSwarmSizeMin=5$, and 50 robots
    are available, we can compute the reallocation rates
    using~\cref{eqn:swarm-availability,eqn:robustness-rho,eqn:robustness-ss-queue-length}
    for the small warehouse, and extrapolate for the medium warehouse as
    needed. }
\item { Minimum collective performance: Achieved by $\TaskedSwarmSizeMin=5$
    robots for all $\kappa$ in the small warehouse, which we extrapolate for the
    medium warehouse, $\TaskedSwarmSizeMin=5*4=20$.}
\item { Robot reliability: Given the tightly controlled indoor environment and
    assumed regular maintenance of robots, the overall robotic temporary
    failure, repair, and permanent failure rates should be negligible, and we
    set $\lambda_{d}=0$ and $\mu_{b}=0$, as $\TheSwarmSize$ is fixed throughout
    operation.}
\end{itemize}
%

We use a $32\times16=512~m^2$ arena with a single-source object distribution for
the small warehouse, and a $48\times48=2,048~m^2$ arena with a random object
distribution for the medium warehouse, modeling smaller and larger scale order
fulfillment operations, respectively.  We use the intuitions
in~\cref{tab:sc1-kappa-summary} and perform a comparative analysis of the swarm
properties, i.e., emergent self-organization, scalability, flexibility, and
robustness of the candidate $\kappa$ algorithms described
in~\cref{ssec:sc-candidate-algs} to determine if our intuitions are supported by
our numerical metric calculations. We omit results where no statistically
significant differences are observed, or where the observed trends are the same
as those for the shown warehouse.
\subsection{Emergent Self-Organization Analysis}\label{ssec:sc1-emergence}
\begin{figure}[t]
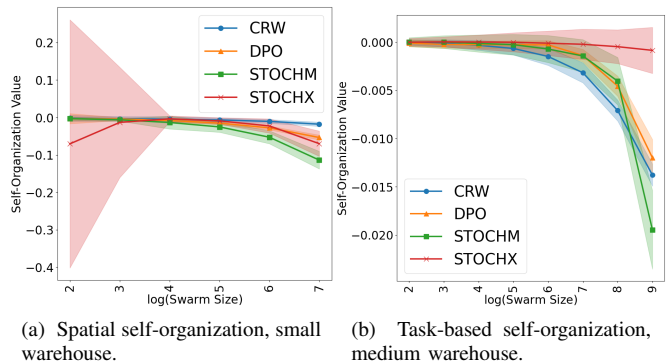

  \captionsetup[subfigure]{position=bottom,width=0.45\linewidth}
  \centering
  \subfloat[\label{fig:sc1-emergence-spatial-small} Spatial self-organization,
  small warehouse.]
  {\includegraphics[width=.49\linewidth]{\figcmproot{sc1}/cc-PM-ss-self-org-ifl-2021trosc1-SS.32x16x2+population_size.Log128.png}}
\subfloat[\label{fig:sc1-emergence-task-medium} Task-based self-organization, medium warehouse.]
  {\includegraphics[width=0.50\linewidth]{\figcmproot{sc1}/cc-PM-ss-self-org-mpg-2021trosc1-RN.48x48x2+population_size.Log512.png}}%
  \caption{\label{fig:sc1-emergence}\footnotesize{Swarm emergent
      self-organization under ideal conditions (no sensor and actuator noise
      or population dynamics). Graphs of similar trends between algorithm
      spatial self-organization in the medium warehouse, no statistically
      significant differences between algorithms in the medium warehouse
      omitted.  }}
\end{figure}

In \cref{fig:sc1-emergence-spatial-small}, the level of spatial
self-organization remains relatively constant across all swarm sizes, indicating
that each $\kappa$ is able to manage increasingly restricted space reasonably
well.  The small arena size restricts the ability of swarms to learn task
parameters by conflating beneficial inter-robot interactions with excessive
collision avoidance, and there are no statistically significant differences in
the spatial self-organization among the more complex STOCHM and STOCHX
algorithms.  We see statistically significant differences between CRW and the
other algorithms in the small warehouse, however, as its purely reactive nature
is relatively unaffected by the conflation.

Spatial self-organization emerges due to sufficiently large perturbations of
robot exploratory random walks via interference by robots returning to the nest
with objects and the otherwise random motion of the swarm for CRW. In more
spatially constrained situations, DPO robots are less random, but compensate for
this with memory. STOCHM and STOCHX robots have additional fluctuations due to
their stochastic task allocation policies. Overall, these observations
demonstrate the importance of the \emph{random fluctuations} system property
necessary for self-organization to occur, as discussed
in~\cref{ssec:meas-emergence-bg}, and that in spatially constrained environments
self-organization can arise even in less ``intelligent'' algorithms due to the
high levels of inter-robot interaction. This suggests that algorithmic
randomness is the most important factor in emergence, together with swarm
density, which was shown in previous work to be critical to develop
self-organization~\cite{Sugawara1997}.

From~\cref{fig:sc1-emergence-task-medium} we see few statistically significant
differences between the level of spatial self-organization for CRW, DPO, and
STOCHM. In this less spatially constrained environment, the swarm must
collectively learn \emph{more} from fewer interactions (i.e., there is less
conflation between different types of inter-robot interactions), demonstrating
the importance of the \emph{multiple interactions} system property necessary for
self-organization to occur, as discussed in~\cref{ssec:meas-emergence-bg}. As
expected, the complex STOCHX algorithm is the most adept at doing this via its
dynamic use of caches, and obtains a near linear scaling in self-organization as
$\TheSwarmSize$ is increased.

\subsection{Scalability Analysis}\label{ssec:sc1-scalability}
\begin{figure}[t]
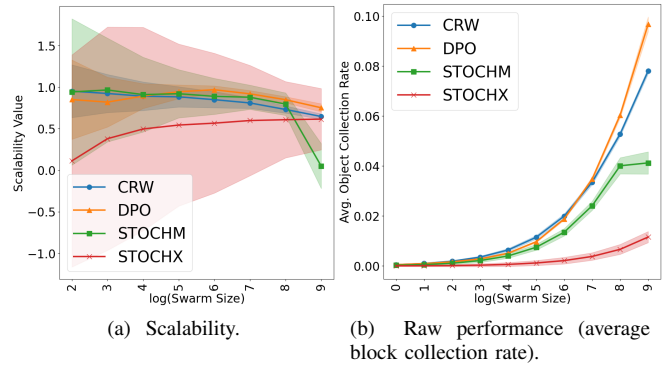

  \captionsetup[subfigure]{position=bottom,width=0.45\linewidth}
  \centering
  \subfloat[\label{fig:sc1-scalability-medium} Scalability.]
  {\includegraphics[width=0.49\linewidth]{\figcmproot{sc1}/cc-PM-ss-scalability-parallel-frac-2021trosc1-RN.48x48x2+population_size.Log512.png}}%
  \subfloat[\label{fig:sc1-perf-medium} Raw performance (average block
  collection rate).]
  {\includegraphics[width=0.49\linewidth]{\figcmproot{sc1}/cc-PM-ss-raw-2021trosc1-RN.48x48x2+population_size.Log512.png}}
  \caption{\label{fig:sc1-scalability-and-perf-medium}\footnotesize{Swarm
      scalability and raw performance under ideal conditions (no sensor and
      actuator noise or population dynamics) in the medium warehouse. Nearly
      identical results for the small warehouse are omitted.}}
\end{figure}

From~\cref{fig:sc1-scalability-and-perf-medium}, we see much tighter confidence
intervals for the simpler CRW and DPO algorithms, and broader intervals for
STOCHM and STOCHX, indicating that while more ``intelligent'' algorithms have
the potential to be highly scalable, their scalability is also more stochastic
than simpler algorithms. This is in agreement with the intuitions laid out
in~\cref{tab:sc1-kappa-summary}: more complex algorithms balance exploring the
solution space with exploiting the current solution, resulting in reduced
performance and scalability, while simpler algorithms perform more exploitation
of the current solution than exploration, resulting in increased performance and
scalability.

Between~\cref{fig:sc1-emergence} and~\cref{fig:sc1-scalability-and-perf-medium}
we see strong orthogonality between self-organization and
scalability/performance: algorithms which are more ``intelligent'' are less
reliably scalable, though they \emph{can} be more scalable than less intelligent
algorithms.  Previous work~\cite{Harwell2020a} has shown that STOCHX outperforms
the other algorithms in foraging scenarios more conducive to the emergence of
strong traffic patterns.  The random foraging environment of the medium
warehouse is much more adverse, leading to the observed lower performance and
scalability. This difference demonstrates the interaction between scenario
characteristics and swarm dynamics, the necessity of accurate modeling and
thorough analysis when designing SR systems, and the importance of matching
algorithms to environments to which they are naturally well-suited.

\subsection{Flexibility Analysis}\label{ssec:sc1-flexibility}
\begin{figure}[t]
\captionsetup[subfigure]{position=bottom,width=0.45\linewidth}
\centering
\subfloat[\label{fig:sc1-reactivity-small}Small warehouse.]
  {\includegraphics[width=.49\linewidth]{\figcmproot{sc1}/cc-PM-ss-reactivity-2021trosc1-SS.32x16x2+temporal_variance.BCSquare.Z50.png}}
  \subfloat[\label{fig:sc1-reactivity-medium} Medium warehouse.]
  {\includegraphics[width=.49\linewidth]{\figcmproot{sc1}/cc-PM-ss-reactivity-2021trosc1-RN.48x48x2+temporal_variance.BCSquare.Z200.png}}%
  \caption{\label{fig:sc1-reactivity}\footnotesize{Swarm reactivity
      $\ReactivityMetric$ across both warehouses under periodic, high intensity
      changes in environmental adversity (square waves). Distance \emph{from}
      ideal environmental conditions is on the X axis, and distance \emph{to}
      the ideal reactivity curve $\OptimalReactivityCurve$ is on the Y axis.
      Lower is better. }}
\end{figure}
\begin{figure}[t]
\captionsetup[subfigure]{position=bottom,width=0.45\linewidth}
\centering
\subfloat[\label{fig:sc1-adaptability-small} Small warehouse.]
  {\includegraphics[width=.49\linewidth]{\figcmproot{sc1}/cc-PM-ss-adaptability-2021trosc1-SS.32x16x2+temporal_variance.BCSquare.Z50.png}}
\subfloat[\label{fig:sc1-adaptability-medium} Medium warehouse.]
  {\includegraphics[width=.49\linewidth]{\figcmproot{sc1}/cc-PM-ss-adaptability-2021trosc1-RN.48x48x2+temporal_variance.BCSquare.Z200.png}}%
  \caption{\label{fig:sc1-adaptability}\footnotesize{Swarm adaptability
      $\AdaptabilityMetric$ across both warehouses under periodic, high
      intensity changes in environmental adversity (square waves). Distance
      \emph{from} ideal environmental conditions is on the X axis, and distance
      \emph{to} the ideal adaptability curve $\OptimalAdaptabilityCurve$ is on
      the Y axis. Lower is better. }}
\end{figure}
In~\cref{fig:sc1-reactivity,fig:sc1-adaptability}, we see more statistically
significant differences between algorithms in reactivity and adaptability for
the medium warehouse than for the small one, which is likely due to the
conflation of the experimental variables for inter-robot interference and
environmental adversity in the small warehouse, as discussed
in~\cref{ssec:sc1-emergence}. We also see further evidence of synergy between
our measures in the correlation between the relative levels of spatial and
task-based self-organization and flexibility.

STOCHM and STOCHX swarms have the highest
levels of self-organization (\cref{fig:sc1-emergence}), and the highest
reactivity and adaptability as well for more disruptive $\NonIdealEnvDeviation$.
The DPO algorithm, while more complex than CRW, is not as reactive or adaptive,
demonstrating that adding cognitive capacity to a given algorithm is not
guaranteed to make the swarm more flexible. DPO swarms are unable to
substantially react to changing environmental conditions because they are
fundamentally unable to alter their behavior, which is compounded with increased
congestion at shared resources, which manifests as a lack of ``intelligence'' in
changing environmental conditions.

STOCHM performs better due to the more complex task allocation distributions,
and STOCHX swarms, as expected, are generally the most reactive and adaptive of
all evaluated algorithms. Both STOCHM and STOCHX are also cognitive, suggesting
that it is possible to design a complex, cognitive algorithm which is just as
flexible as a reactive algorithm---an important insight during the design
process. More work is needed to further explore this relationship, but our
results suggest that there is a threshold of emergent self-organization for
cognitive algorithms which must be met in order to obtain collective performance
which consistently outperforms that of simpler biomimetic algorithms.

In the medium warehouse
(\cref{fig:sc1-reactivity-medium,fig:sc1-adaptability-medium}), because the
relative levels of inter-robot interference are lower, we see much more
consistent trendlines of approximately the same slope for STOCHX swarms,
indicating that such swarms are sufficiently ``intelligent'' to maintain nearly
identical marginal reactivity to increasingly adverse environmental conditions
(i.e., very small slopes). Other algorithms, having less ``intelligence,'' show
decreasing marginal reactivity (i.e., increasing slopes).

\subsection{Robustness Analysis}\label{ssec:sc1-robustness}
\begin{figure}[t]
\captionsetup[subfigure]{position=bottom,width=0.45\linewidth}
\centering
\subfloat[\label{fig:sc1-robustness-sa-small} Small warehouse.]
  {\includegraphics[width=.49\linewidth]{\figcmproot{sc1}/cc-PM-ss-robustness-saa-2021trosc1-SS.32x16x2+saa_noise.all.C8.Z50.png}}
  \subfloat[\label{fig:sc1-robustness-sa-medium} Medium warehouse.]
  {\includegraphics[width=.49\linewidth]{\figcmproot{sc1}/cc-PM-ss-robustness-saa-2021trosc1-RN.48x48x2+saa_noise.all.C8.Z200.png}}

  \caption{\label{fig:sc1-robustness-sa}\footnotesize{Swarm robustness to sensor and actuator
      noise ($\SARobustnessMetric$). Standard deviation $\sigma$ of
      sensor readings/actuator effects from their actual/intended values is on
      the X-axis, distance from the ideal robustness curve
      $\PerfCurveSub{\TheSwarmSize}{ideal}$ is on the Y axis. Lower is better.}}
\end{figure}

For the small warehouse in~\cref{fig:sc1-robustness-sa-small}, the CRW and
STOCHX algorithms generally show statistically equivalent robustness. Of all the
candidate algorithms, CRW is the least reliant on the accuracy of
sensors/actuators of all algorithms, while STOCHX has a high level of
self-organization---both equivalent means of achieving robustness to sensor and
actuator noise. DPO and STOCHM likewise are not statistically different in terms
of sensor and actuator noise robustness, except at higher $\sigma$.  We see a
clearer separation between algorithms in the medium warehouse
in~\cref{fig:sc1-robustness-sa-medium}, again likely due to the reduced
conflation between different types of inter-robot interactions.  In
both~\cref{fig:sc1-robustness-sa-small,fig:sc1-robustness-sa-medium}, we also
see strong evidence of ``phase transitions'' of the behavior of all algorithms
with $\sigma \ge 0.017$. The transition is much more pronounced for DPO and
STOCHM, in which the ability of these algorithms to handle injected noise begins
to break down---a crucial insight in the design process.
\begin{figure}[t]
  \subfloat[\label{fig:sc1-availability-small} $L=\{5,10,20\}$.]
      {\includegraphics[width=0.49\linewidth]{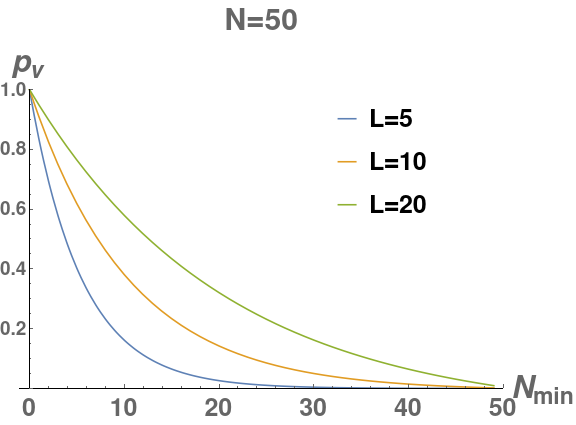}}
  \subfloat[\label{fig:sc1-availability-medium} $L=\{5,10,20\}$. ]
      {\includegraphics[width=0.49\linewidth]{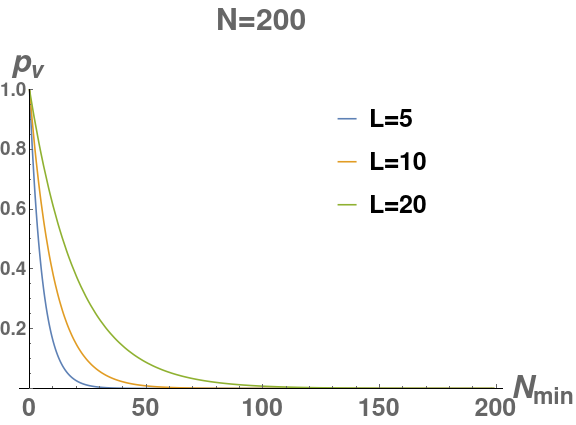}}
      \caption{\footnotesize{Swarm availability $p_{v}$ for desired steady state
          swarm sizes $\TaskedSwarmSize{t}$ (queue lengths $L$) for different
          $\TaskedSwarmSizeMin$, the minimum number of robots required to meet a
          desired performance level, given Poisson-distributed reallocations of
          robots in the swarm $\TheSwarm$ to other tasks.}}
\end{figure}
\begin{figure}[t]
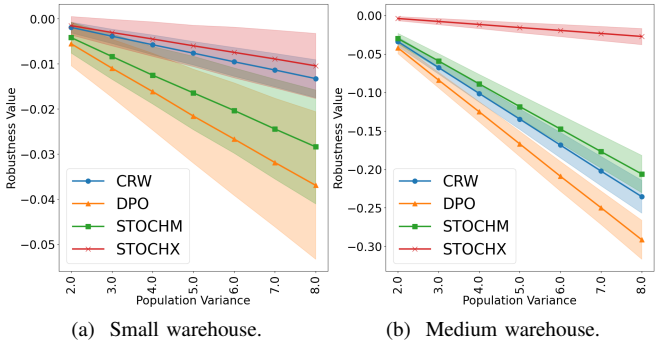

\captionsetup[subfigure]{position=bottom,width=0.45\linewidth}
\centering
\subfloat[\label{fig:sc1-robustness-small-pd} Small warehouse. ]
{\includegraphics[width=.49\linewidth]{\figcmproot{sc1}/cc-PM-ss-robustness-pd-2021trosc1-SS.32x16x2+population_dynamics.C8.F1p0.M0p001.R0p001022222.png}}%
\subfloat[\label{fig:sc1-robustness-medium-pd} Medium warehouse.]
{\includegraphics[width=.49\linewidth]{\figcmproot{sc1}/cc-PM-ss-robustness-pd-2021trosc1-RN.48x48x2+population_dynamics.C8.F1p0.M0p001.R0p001005556.png}}
\caption{\label{fig:sc1-robustness-pd}\footnotesize{ Swarm population dynamics
    robustness ($\PDRobustnessMetric$). Population variance compared to ideal
    conditions, i.e., how much \emph{more} often robots from $\TheSwarm$ are
    re-allocated to other tasks, is on the X-axis, and the computed robustness
    value on the Y-axis. }}
\end{figure}
%

To analyze population dynamics robustness, we use $\TheSwarmSize=50$ for the
small warehouse, per the company's specification,
and~\cref{eqn:swarm-availability} to solve for $p_v$, given desired steady state
$\TaskedSwarmSize{t}=\{5,10,20\}$, shown in~\cref{fig:sc1-availability-small}.
Given that sufficient performance can be achieved (per scenario specification)
with $\TaskedSwarmSizeMin=5$ robots for any algorithm, we see that this
corresponds to $p_v\sim{}0.40$, meaning that approximately 40\% of the time all
5 robots of a given team were available to work on filling an order.  For the
medium warehouse, we use $\TheSwarmSize=200$ and $\TaskedSwarmSizeMin=20$,
extrapolating from the company's specification, and solve for $p_v$ in the same
manner, shown in~\cref{fig:sc1-availability-medium}. With $\TheSwarmSize=200$,
the availability curves drop off much more steeply, and we must have larger team
sizes to achieve the same level of availability as the small warehouse.

We again see the same dichotomy between STOCHM/DPO and STOCHX/CRW in the small
warehouse in~\cref{fig:sc1-robustness-small-pd} for higher population variances:
the emergent self-organization of the STOCHX algorithm makes it the most robust
to fluctuating swarm sizes, while other algorithms struggle. We speculate that
the dynamic use of caches by STOCHX  allows additional information about
the current swarm task distribution to be encoded into the environment, which
can then be partially recovered by returning/new robots, per our intuition.
In~\cref{fig:sc1-robustness-medium-pd}, we see much stronger evidence of synergy
between our metrics: the relative ordering mirrors the level of emergent
self-organization, with the caveat of a possible lack of statistically
significant differences between STOCHM and CRW.

We have discussed swarm self-organization, scalability, flexibility, and
robustness for the warehouse scenario for all $\kappa$. It would now be up to the
company to take our results and determine which of the four main
swarm properties are most important to them to  make a final
algorithm decision.

Overall, we see strong evidence in this section's analyses of the
synergistic nature of our metrics for SR system properties: trends within
and between curves on graphs of each can be intuitively explained, and easily
mapped back to the technical specifications of a given method in ways that are
not evident from raw performance curves. We have observed strong correlations
between emergent self-organization, scalability, flexibility, and robustness, and our
intuitions in~\cref{tab:sc1-kappa-summary}.

\section{Scenario 2: Outdoor Search and Rescue}\label{sec:sc2}
This scenario is inspired by the human-swarm interaction discussion
in~\cite{CarrilloZapata2020}, in which we as SR system designers are tasked with
choosing an appropriate control algorithm for an outdoor search and rescue
task. The system will be deployed in disaster zones/military conflict zones in a
fully autonomous fashion as a first response measure, to assess a given outdoor
environment and to search for ``objects'' of interest within the
environment. These objects can be unconscious civilians trapped in rubble, or
places of utility line breaks (e.g., gas lines), which need to be quickly
discovered and reported to human responders for further investigation in order
to minimize loss of life and further infrastructure damage. Some civilians may
be conscious and trying to make their way out of the disaster zone or help
others, in which case they need to be found and guided to the nearest safe
zone. Similarly, the location of utility line breaks may appear to ``move'' as
robot onboard sensing is not powerful enough to provide precise coordinates in
potentially occluding environmental conditions.

Moderate levels of robot losses are expected due to the instability of the
environment, despite hardened robotic hardware. Therefore, relatively large
$\TheSwarmSize$ is required to ensure the success of the mission; we evaluate
all algorithms with $\TheSwarmSize=\{12,115,320\}$ (12 used for baseline
reference), and show raw performance results with
$\TheSwarmSize=12\ldots820$. The selected control algorithm needs to be highly
scalable and intelligent, so that it can efficiently adapt to a wide range of
problem scales and unknown environmental conditions, and provide stable
performance across highly volatile environmental conditions.

\begin{table}[t]
  \centering\scriptsize
  \begin{tabularx}{\linewidth}{@{}p{.75cm}YY@{}}
    $\kappa$ & Intuitions\\

    \toprule
    CRW &
          \begin{minipage}[t]{\linewidth}
            \begin{itemize}
            \item {No negative effects from moving targets, strong robustness to
                fluctuating populations, and high flexibility because of lack of
                memory. }
            \item { Robustness to sensor noise because of minimal sensor usage. }
            \end{itemize}
          \end{minipage}\\

    \midrule
    DPO &

          \begin{minipage}[t]{\linewidth}
            \begin{itemize}
            \item {Lower levels of self-organization than CRW, because its world
                model assumes static object locations.}
            \item {Low robustness to sensor and actuator noise because of high
                sensor usage. }
            \item {Minimal population dynamics robustness, due to dependence on
                accurate world models for efficiency.}
            \item { Low flexibility due to lack of ability to change allocated
                task.}
            \end{itemize}
          \end{minipage}\\

    \midrule
    STOCHM &

             \begin{minipage}[t]{\linewidth}
               \begin{itemize}
               \item {Poor self-organization and population dynamics robustness
                   due to static cache usage with moving targets; minimal
                   information about task distribution can be encoded into
                   caches because they are maintained by an outside process. }
               \item {Low population dynamics robustness due to dependence of
                   behavior on accurate robot world models, but higher than DPO,
                   because some information about the world model carried by
                   robots is now encoded in the environment, and the ability to
                   choose from multiple tasks should also improve collective
                   re-convergence to steady-state.}
               \end{itemize}
             \end{minipage}\\

    \midrule
    STOCHX &
             \begin{minipage}[t]{\linewidth}
               \begin{itemize}
               \item {Modest self-organization due to dynamic cache usage and
                   highly flexible task allocation distribution despite moving
                   targets; better than STOCHM because caches can be created and
                   depleted to help offset target motion. }
               \item {Modest population dynamics robustness, flexibility due to the
                   ability to encode the current task distribution into environmental
                   state for inexperienced robots to find via caches (or a lack
                   thereof). }
               \end{itemize}
             \end{minipage}\\
    \bottomrule

  \end{tabularx}\caption{\footnotesize{Summary of control algorithm ($\kappa$) intuitions, along
      with comparative strengths/weaknesses in the outdoor search and
      rescue scenario.}}\label{tab:sc2-kappa-summary}
\end{table}
\subsection{Design Constraints}\hfill

This scenario, or variants of it, has been studied
by~\cite{Hsieh2008,Kumar2003}.  We define the following design constraints:
\begin{itemize}
\item {$\PerfCurve{\TheSwarmSize}$: measures the rate at which objects are picked
    up for the first time; more efficient $\kappa$ will discover more objects faster.}
\item {Operating arena: Variable size with constant swarm density; size of the arena
    in which targets can be found
    is not known beforehand. Given the limited sensing
    capabilities of the robots, and the anticipated robot losses due to
    malfunction or environmental issues, relatively large swarm sizes should be
    tested.  }
\item {Object distribution: Not known \emph{a priori}, so random or power law
    are appropriate models; we choose power law. Objects can move randomly over
    time; it is not known how often this happens so we evaluate performance with
    $p_{rw}=0-0.051$.}
\item {Environmental disturbances: Potentially large gradually evolving
    environmental disturbances during swarm operation, such that the swarm
    operating conditions will change from favorable to unfavorable or vice
    versa, and back again throughout swarm operation (e.g., a passing storm,
    windy conditions, change in visibility, etc).  We apply throttling functions
    to the maximum robot speed during operation, and use a sinusoidal
    disturbance model with throttling amplitudes $0-0.4$ and frequency
    $f=10000$, modeling cyclically shifting environmental conditions roughly
    every 2.75 hours. }
\item {Sensor and actuator noise: Environmental disturbances may occur, such as
    clouds interfering with localization, fluctuating lighting conditions
    interfering with perception, etc. Given our hardened robotic hardware, we
    select a Gaussian noise model $G(\mu,\sigma)$ with $\mu=0$ as an appropriate
    model, and test with $\sigma=0-0.1$.}
\item {Robot reliability: Environmental disturbances and the general
    unpredictability of operating in an outdoor environment may give rise to
    software or hardware artifacts which will cause robots to fail permanently,
    so we test $\lambda_{d}=0-0.0007$.}
\item {Task reallocations: No unexpected task reallocations, due to the time
    sensitivity of the application, so we set $\lambda_{bd}=\mu_{bd}=0$.}
\end{itemize}
We develop the intuitions shown in~\cref{tab:sc2-kappa-summary}, and perform a
comparative analysis of the emergent self-organization, scalability,
flexibility, and robustness of the candidate $\kappa$ described
in~\cref{ssec:sc-candidate-algs} to determine if our intuitions are supported by
our numerical measurements. We omit results where no statistically significant
differences are observed.

\subsection{Emergent Self-Organization Analysis}\label{ssec:sc2-emergence}\hfill
\begin{figure}[t]
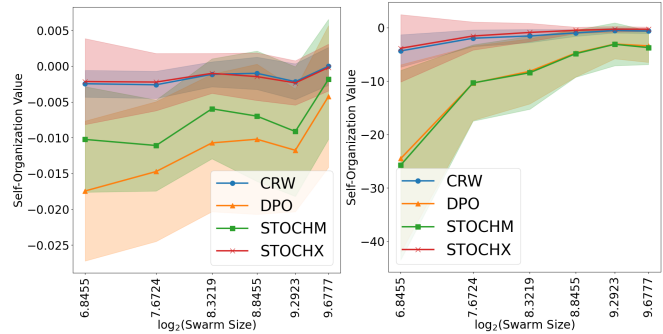

  \captionsetup[subfigure]{position=bottom,width=0.45\linewidth}
  \centering
  \subfloat[Task-based self-organization, no target motion.]
  {\includegraphics[width=.50\linewidth]{\figcmproot{sc2}/cc-PM-ss-self-org-mpg-2021trosc2-PL.16x16x2+block_motion_dynamics.C4.F25p0.RW0p001+population_constant_density.5p0.I16.C8_0.png}}
  \subfloat[Spatial self-organization, maximum target motion.]
  {\includegraphics[width=.48\linewidth]{\figcmproot{sc2}/cc-PM-ss-self-org-mfl-2021trosc2-PL.16x16x2+block_motion_dynamics.C4.F25p0.RW0p001+population_constant_density.5p0.I16.C8_1.png}}
  \caption{\label{fig:sc2-emergence} \footnotesize{Swarm emergent
      self-organization with constant swarm density under ideal conditions (no
      sensor and actuator noise or population dynamics).}}
\end{figure}

In~\cref{fig:sc2-emergence}, we see a strong dichotomy: STOCHX and CRW swarms
show consistent levels of self-organization across swarm sizes for both high and
low rates of target motion, while the self-organization of DPO and STOCHM swarms
is much more sensitive to swarm sizes across both target motion rates. As swarm
sizes increase, it naturally becomes easier to find moving targets through sheer
numbers rather than intelligent algorithm design. At smaller swarm sizes, STOCHX
exhibited the strongest trends (marginally), and was therefore able to overcome
some of the world model errors introduced by target motion through the emergence
of specialized task allocation distributions, dynamic cache usage, and bucket
brigading, while the (partially) random walk used in CRW also proves an
effective strategy, despite its simplicity.

DPO relies heavily on static target locations, while STOCHM relies on both
static target and static cache locations. As a result, much of the learning of
STOCHM swarms to utilize caches is wasted due to the unconstrained nature of the
environment and target motion. Overall, the strategies used by these algorithms
are much less effective when compared with STOCHX and CRW, as we would
intuitively expect.

\begin{figure}[t]
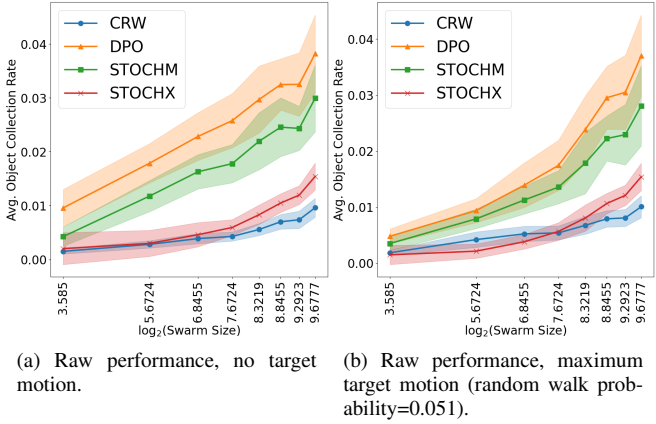

\captionsetup[subfigure]{position=bottom,width=0.45\linewidth}
\centering
\subfloat[Raw performance, no target motion.]
  {\includegraphics[width=.49\linewidth]{\figcmproot{sc2}/cc-PM-ss-raw-2021trosc2-PL.16x16x2+block_motion_dynamics.C4.F25p0.RW0p001+population_constant_density.5p0.I16.C8_0.png}}
  \subfloat[Raw performance, maximum target motion (random walk
  probability=0.051).]
  {\includegraphics[width=.49\linewidth]{\figcmproot{sc2}/cc-PM-ss-raw-2021trosc2-PL.16x16x2+block_motion_dynamics.C4.F25p0.RW0p001+population_constant_density.5p0.I16.C8_3.png}}
  \caption{\label{fig:sc2-raw-perf} \footnotesize{Swarm raw performance (rate of
      object pickup), mimicking discovery of moving targets in the scenario for
      $\TheSwarmSize=12\ldots820$. }}
\end{figure}
From~\cref{fig:sc2-raw-perf}, we observe a negative correlation between emergent
self-organization and performance, in contrast to~\cite{Harwell2020a}. This
suggests that the correlation between self-organization and performance is
highly dependent on the nature of the environment and that in dynamic or
extremely adversarial environments, less ``intelligent'' algorithms may perform
better, as we would expect.

\subsection{Scalability Analysis}\label{ssec:sc2-scalability}
The extremely adverse power law object distribution in this scenario is not
conducive to self-organization because of its irregular block clusters. In
combination with a constant swarm density and target motion this leads to no
statistically significant differences in scalability between the
algorithms. This result is in contrast to~\cite{Harwell2020a}, in which
differences in emergent self-organization were positively correlated with
statistically significant differences in scalability. This again suggests that
positive correlation only holds in static environments, and that in dynamic and
adversarial environments, ``intelligence'' plays a minimal role in determining
scalability.


Having analyzed the effect of target movement rate, we set a moderate rate of
$0.001$ to simplify the rest of our analyses.
\subsection{Flexibility Analysis}\label{ssec:sc2-flexibility}
\begin{figure}[t]
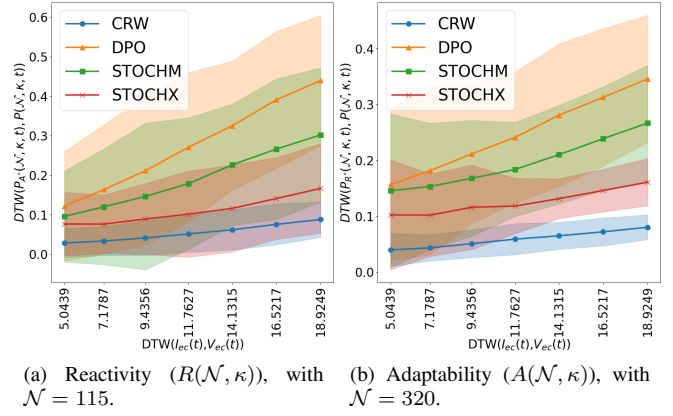

\captionsetup[subfigure]{position=bottom,width=0.45\linewidth}
\centering
\subfloat[\label{fig:sc2-flexibility-reactivity}Reactivity
  ($\ReactivityMetric$), with $\TheSwarmSize=115$.]
  {\includegraphics[width=.49\linewidth]{\figcmproot{sc2}/cc-PM-ss-adaptability-2021trosc2-PL.16x16x2+population_constant_density.5p0.I32.C3+temporal_variance.MSine_1.png}}
  \subfloat[\label{fig:sc2-flexibility-adaptability}Adaptability
  ($\AdaptabilityMetric$), with $\TheSwarmSize=320$.]
  {\includegraphics[width=.49\linewidth]{\figcmproot{sc2}/cc-PM-ss-reactivity-2021trosc2-PL.16x16x2+population_constant_density.5p0.I32.C3+temporal_variance.MSine_2.png}}
  \caption{\label{fig:sc2-flexibility} \footnotesize{Swarm flexibility with
      constant swarm density under sinusoidal fluctuations in environmental
      adversity. Distance \emph{from} ideal environmental conditions is on the X
      axis, and distance \emph{to} the ideal adaptability/reactivity curve is on
      the Y axis. Lower is better.} The graphs of reactivity with
    $\TheSwarmSize=320$ and adaptability with $\TheSwarmSize=115$ are
    statistically identical to the shown graphs, and are omitted.}
\end{figure}
In~\cref{fig:sc2-flexibility-reactivity}, we see the same dichotomy observed
in~\cref{ssec:sc2-emergence,ssec:sc2-scalability}: CRW and STOCHX are the most
reactive algorithms, and DPO and STOCHM are the least. However, this dichotomy
only emerges for higher levels of environmental disturbances; for smaller
levels, the algorithms are largely statistically equivalent. Overall, as
expected, the CRW and STOCHX algorithms are the most reactive, due to their
memory-less nature and ability to dynamically utilize caches, respectively. DPO
and STOCHM, which rely on static object and cache locations, are much less
reactive.

In~\cref{fig:sc2-flexibility-adaptability}, we see fewer significant differences
between algorithm adaptability, regardless of the level of environmental
disturbance, with the exception of CRW showing significant differences with DPO
and STOCHM for higher levels of environmental disturbance. In the presence of
randomized target motion, we expect it to be difficult for algorithms to
\emph{also} adapt to changing environmental conditions. In this instance, the
memory-less CRW algorithm is the only algorithm able to statistically
differentiate itself from the other evaluated algorithms.

\subsection{Robustness Analysis}\label{ssec:sc2-robustness}\hfill
\begin{figure}[t]
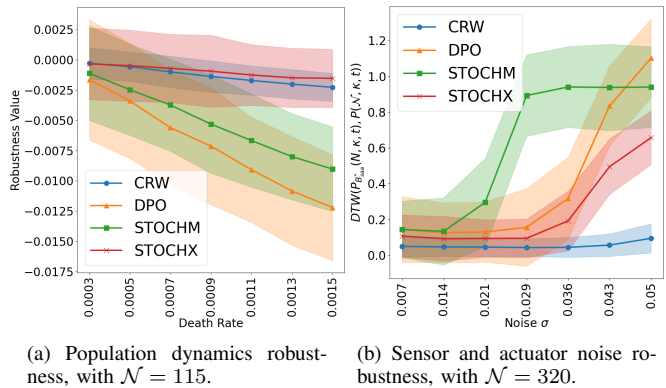

\captionsetup[subfigure]{position=bottom,width=0.45\linewidth}
\centering
\subfloat[\label{fig:sc2-robustness-pd}Population dynamics robustness, with $\TheSwarmSize=115$.]
  {\includegraphics[width=.51\linewidth]{\figcmproot{sc2}/cc-PM-ss-robustness-pd-2021trosc2-PL.16x16x2+population_constant_density.5p0.I32.C3+population_dynamics.C8.F2p0.D0p0001_1.png}}
  \subfloat[\label{fig:sc2-robustness-sa}Sensor and actuator noise robustness, with $\TheSwarmSize=320$.]
  {\includegraphics[width=.48\linewidth]{\figcmproot{sc2}/cc-PM-ss-robustness-saa-2021trosc2-PL.16x16x2+population_constant_density.5p0.I32.C3+saa_noise.all.C8_2.png}}
  \caption{\label{fig:sc2-robustness} \footnotesize{Swarm robustness with
      constant swarm density. In~\cref{fig:sc2-robustness-sa} swarm sensor and
      actuator noise robustness is shown under non-ideal sensors and actuators
      with Gaussian-distributed responses. $\sigma$ of applied noise is on the X
      axis, and distance \emph{to} the ideal sensor and actuator noise
      robustness curve $\PerfCurveSub{\TheSwarmSize}{ideal}$ is on the Y axis.
      Lower is better. In~\cref{fig:sc2-robustness-pd}, swarm population
      dynamics robustness ($\PDRobustnessMetric$) is shown under pure death
      dynamics.  The graphs of sensor and actuator noise robustness with
      $\TheSwarmSize=115$ and population dynamics robustness with
      $\TheSwarmSize=320$ are statistically identical to the shown graphs, and
      are omitted.
    }}
\end{figure}

In~\cref{fig:sc2-robustness-sa}, we see identical trends for each individual
$\kappa$, as well as identical trends between $\kappa$. The robustness of CRW
and STOCHX swarms to sensor and actuator noise is statistically identical in
most cases for increasing $\sigma$. For CRW swarms this is likely due to their
low reliance on the accuracy of sensors and actuators during operation: the more
reliant a swarm control algorithm is on accurate sensor and actuator
information, the less robust it will be to adverse environmental conditions
which negatively affect sensors and actuators. For STOCHX swarms this is likely
due to their ability to be flexible and switch tasks dynamically to attempt to
mitigate the noise (e.g., less transporting of objects over long distances,
which is more prone to localization errors).

DPO, STOCHM, and STOCHX swarms have noticeably lower robustness for larger
values of $\sigma$, and undergo phase changes at $\sim{\sigma=0.01-0.05}$, with
the overall robustness dramatically dropping for larger $\sigma$. This is the
inflection point at which the combination of increasing noise and moving objects
overwhelms the swarm's cognitive ability to compensate; it occurs at much lower
$\sigma$ for STOCHM swarms than those for DPO and STOCHX, due to their usage of
static caches as they attempt to adapt.

In~\cref{fig:sc2-robustness-pd}, CRW swarms exhibit a high level of population
dynamics robustness, due to their memory-less nature; similarly for STOCHX
swarms with their ability to encode the current task distribution into
environmental state for inexperienced robots to find via caches. DPO and STOCHM,
lacking such abilities, have much lower robustness. For the omitted graph with
$\TheSwarmSize=320$, statistically identical responses to pure death dynamics
follow directly from our intuition: with large swarm sizes, the loss of an
individual robot affects the swarm's overall progress on its task much
less. However, this observation is likely tied to the extremely adversarial
nature of the scenario, and we would reasonably expect differences to emerge in
a more favorable environment.

Our results in this section suggest strong synergy between our robustness and
flexibility measures, similar to what was observed for the first scenario. We
have again seen quantitative confirmation of our intuitions for each $\kappa$
through our derived metrics for each of the swarm properties of
self-organization, scalability, flexibility, and robustness. It would now be up
to project managers to take our recommendations and make a final
decision. However, our results suggest that the STOCHX algorithm is, on the
whole, best suited to the outdoor search and rescue application.

%
\section{Discussion}\label{sec:discussion}

The application of our proposed metrics to the object transport and
search-and-rescue scenarios in~\cref{sec:sc1,sec:sc2} provides a weakly
inductive proof of their correctness and utility within the foraging domain, if
not more broadly. Despite fundamental differences in characteristics between
scenarios, our metrics were still able to expose the same underlying traits of
each controller which originated in our intuitions
(\cref{tab:sc1-kappa-summary,tab:sc2-kappa-summary}). We saw similar dichotomies
emerge between the CRW/STOCHX and DPO/STOCHM controllers, and that while more
``intelligent'' algorithms might not be the most scalable or highly performing,
depending on environmental conditions and scenario characteristics, they were
reliably more flexible and robust. This is an important insight for swarm
engineers to consider when designing SR systems, and is not easily obtainable
from simply comparing raw performance results.

Furthermore, our methodology provided a clear mapping from results back to our
intuitions for the evaluated algorithms in both scenarios. We therefore argue
that our proposed metrics are insightful tools for making theoretically grounded
decisions on algorithm selection from high level algorithm descriptions, such as
those we presented in~\cref{sec:sc-overview}, for many real-world applications
beyond the foraging tasks discussed here, and they should be included in the
swarm engineer's toolbox.

Our general definition of $\PerfCurve{\TheSwarmSize}$ is an arbitrary
performance measure, and by basing all of our measures on it, it is easy to
analyze other foraging algorithms. For example, in analyzing a partial
differential equation behavioral prediction approach to foraging, we could
define $\PerfCurve{\TheSwarmSize}$ as the number of robots carrying objects back
to the nest~\cite{Lerman2001}, or how well population fractions engaged on
various tasks are balanced~\cite{Lerman2006}. In~\cite{Lu2020}, the use
of dynamic depots could lead to a logical definition of
$\PerfCurve{\TheSwarmSize}$ as the average load level or wait time of each
depot.

We emphasize again that the definitions of our measures were intended to be
application agnostic; it is possible to define more precise definitions specific
to a given application. For example, one could define emergent self-organization
according to the rate at which something happens (e.g., rate of construction
progress~\cite{Petersen2011}), with the rate of a baseline algorithm with a
``random'' construction strategy subtracted. Such a definition would have more
utility in the realm of autonomous construction than our definition, but would
have little outside of it; our measures, while not as precise as
application-specific ones, can be used broadly across many sub-fields in swarm
robotics.

\section{Conclusions and Future Work}\label{sec:conclusion}

We have presented a set of metrics for more precise characterization of
swarm emergence, scalability, flexibility, and robustness in order to provide
mathematical tools to support the iterative design process of SR systems within
swarm engineering. We demonstrated the utility of our methodology in the context
of two different foraging tasks under various design constraints, and have shown
that they provide intuitive insight into recommending a given method for a
scenario.

Next steps in this research include applying our methodology to other domains
within SR, such as collective transport, pattern formation, etc. We are also
interested in combining our derived measures with predictive modeling and
investigating how analytical derivations of $\PerfCurve{\TheSwarmSize}$ and
$\TLost{\TheSwarmSize}$ from scenario input parameters can be leveraged to
predict swarm emergent self-organization, scalability, flexibility, and
robustness from fundamental principles, and have begun investigating this in our
ongoing work.

In order to facilitate future research and collaboration, the code used for this
research is open source, and can be found at
https://github.com/swarm-robotics/fordyca and
https://github.com/swarm-robotics/sierra.
%

\bibliographystyle{IEEEtran}
\bibliography{2021-TRO}

\begin{thebibliography}{10}
\providecommand{\url}[1]{#1}
\csname url@samestyle\endcsname
\providecommand{\newblock}{\relax}
\providecommand{\bibinfo}[2]{#2}
\providecommand{\BIBentrySTDinterwordspacing}{\spaceskip=0pt\relax}
\providecommand{\BIBentryALTinterwordstretchfactor}{4}
\providecommand{\BIBentryALTinterwordspacing}{\spaceskip=\fontdimen2\font plus
\BIBentryALTinterwordstretchfactor\fontdimen3\font minus
  \fontdimen4\font\relax}
\providecommand{\BIBforeignlanguage}[2]{{%
\expandafter\ifx\csname l@#1\endcsname\relax
\typeout{** WARNING: IEEEtran.bst: No hyphenation pattern has been}%
\typeout{** loaded for the language `#1'. Using the pattern for}%
\typeout{** the default language instead.}%
\else
\language=\csname l@#1\endcsname
\fi
#2}}
\providecommand{\BIBdecl}{\relax}
\BIBdecl

\bibitem{Sahin2005}
E.~{\c{S}}ahin, ``Swarm robotics: From sources of inspiration to domains of
  application,'' in \emph{Swarm Robotics}, ser. LNCS 3342.\hskip 1em plus 0.5em
  minus 0.4em\relax Springer, 2005, pp. 10--20.

\bibitem{Dorigo2013}
M.~{Dorigo} \emph{et~al.}, ``Swarmanoid: A novel concept for the study of
  heterogeneous robotic swarms,'' \emph{IEEE Robotics Automation Magazine},
  vol.~20, no.~4, pp. 60--71, 2013.

\bibitem{Rizk2019}
Y.~Rizk, M.~Awad, and E.~W. Tunstel, ``Cooperative heterogeneous multi-robot
  systems: A survey,'' \emph{ACM Comput. Surv.}, vol.~52, no.~2, Apr. 2019.

\bibitem{Ramachandran2020}
R.~K. {Ramachandran}, N.~{Fronda}, and G.~S. {Sukhatme}, ``Resilience in
  multi-robot target tracking through reconfiguration,'' in \emph{2020 IEEE
  International Conference on Robotics and Automation (ICRA)}, 2020, pp.
  4551--4557.

\bibitem{Labella2006}
T.~H. Labella and M.~Dorigo, ``Division of labor in a group of robots inspired
  by ants' foraging behavior,'' \emph{ACM Trans. on Autonomous and Adaptive
  Systems (TAAS)}, vol.~1, no.~1, pp. 4--25, Sep. 2006.

\bibitem{Hecker2015}
J.~P. Hecker and M.~E. Moses, ``{Beyond pheromones: evolving error-tolerant,
  flexible, and scalable ant-inspired robot swarms},'' \emph{Swarm
  Intelligence}, vol.~9, no.~1, pp. 43--70, 2015.

\bibitem{Kumar2003}
V.~Kumar and F.~Sahin, ``{Cognitive maps in swarm robots for the mine detection
  application},'' \emph{Proc. IEEE Int'l Conference on Systems, Man and
  Cybernetics}, vol.~4, pp. 3364--3369, 2003.

\bibitem{CarrilloZapata2020}
D.~Carrillo-Zapata, E.~Milner, J.~Hird, G.~Tzoumas, P.~J. Vardanega,
  M.~Sooriyabandara, M.~Giuliani, A.~F.~T. Winfield, and S.~Hauert, ``Mutual
  shaping in swarm robotics: User studies in fire and rescue, storage
  organization, and bridge inspection,'' \emph{Frontiers in Robotics and AI},
  vol.~7, p.~53, 2020.

\bibitem{Castello2016}
E.~Castello, T.~Yamamoto, F.~D. Libera, W.~Liu, A.~F. Winfield, Y.~Nakamura,
  and H.~Ishiguro, ``Adaptive foraging for simulated and real robotic swarms:
  the dynamical response threshold approach,'' \emph{Swarm Intelligence},
  vol.~10, no.~1, pp. 1--31, 2016.

\bibitem{Winfield2005a}
A.~F.~T. Winfield and J.~Sa, ``On formal specification of emergent behaviours
  in swarm robotic systems,'' \emph{Science}, pp. 363--371, 2005.

\bibitem{Galstyan2005}
A.~Galstyan, T.~Hogg, and K.~Lerman, ``Modeling and mathematical analysis of
  swarms of microscopic robots,'' in \emph{Proceedings 2005 IEEE Swarm
  Intelligence Symposium, 2005. SIS 2005.}, 2005, pp. 201--208.

\bibitem{Hunt2020}
E.~R. Hunt, ``Phenotypic plasticity provides a bioinspiration framework for
  minimal field swarm robotics,'' \emph{Frontiers in Robotics and AI}, vol.~7,
  p.~23, 2020.

\bibitem{DeWolf2005}
T.~De~Wolf and T.~Holvoet, ``Emergence versus self-organisation: Different
  concepts but promising when combined,'' in \emph{Engineering Self-Organising
  Systems}, S.~A. Brueckner, G.~Di~Marzo~Serugendo, A.~Karageorgos, and
  R.~Nagpal, Eds.\hskip 1em plus 0.5em minus 0.4em\relax Springer Berlin
  Heidelberg, 2005, pp. 1--15.

\bibitem{Szabo2014}
C.~Szabo, Y.~M. Teo, and G.~K. Chengleput, ``Understanding complex systems:
  Using interaction as a measure of emergence,'' in \emph{Proc. Winter
  Simulation Conference}, Dec 2014, pp. 207--218.

\bibitem{Cotsaftis2009}
M.~Cotsaftis, ``An emergence principle for complex systems,'' in \emph{Complex
  Sciences}.\hskip 1em plus 0.5em minus 0.4em\relax Springer Berlin Heidelberg,
  2009, pp. 1105--1117.

\bibitem{Matthey2009}
L.~Matthey, S.~Berman, and V.~Kumar, ``Stochastic strategies for a swarm
  robotic assembly system,'' \emph{Proc. IEEE Int'l Conf. on Robotics and
  Automation}, pp. 1953--1958, 2009.

\bibitem{Agassounon2001}
W.~{Agassounon}, A.~{Martinoli}, and R.~{Goodman}, ``A scalable, distributed
  algorithm for allocating workers in embedded systems,'' in \emph{2001 IEEE
  International Conference on Systems, Man and Cybernetics. e-Systems and e-Man
  for Cybernetics in Cyberspace (Cat.No.01CH37236)}, vol.~5, 2001, pp.
  3367--3373 vol.5.

\bibitem{Lerman2006}
K.~Lerman, C.~Jones, A.~Galstyan, and M.~J. Matar{\'{i}}c, ``{Analysis of
  dynamic task allocation in multi-robot systems},'' \emph{International
  Journal of Robotics Research}, vol.~25, no.~3, pp. 225--241, 2006.

\bibitem{Lu2020}
Q.~{Lu}, G.~M. {Fricke}, T.~{Tsuno}, and M.~E. {Moses}, ``A bio-inspired
  transportation network for scalable swarm foraging,'' in \emph{Proc. IEEE
  Int'l Conf. on Robotics and Automation}, 2020, pp. 6120--6126.

\bibitem{Harwell2019a}
J.~Harwell and M.~Gini, ``Swarm engineering through quantitative measurement of
  swarm robotic principles in a 10,000 robot swarm,'' in \emph{Proc.
  Twenty-Eighth Int'l Joint Conference on Artificial Intelligence, {IJCAI-19}},
  Aug. 2019, pp. 336--342.

\bibitem{Just2017}
W.~A. {Just} and M.~E. {Moses}, ``Flexibility through autonomous
  decision-making in robot swarms,'' in \emph{2017 IEEE Symposium Series on
  Computational Intelligence (SSCI)}, 2017, pp. 1--8.

\bibitem{Winfield2008}
A.~F.~T. Winfield, W.~Liu, J.~Nembrini, and A.~Martinoli, ``Modelling a
  wireless connected swarm of mobile robots,'' \emph{Swarm Intelligence},
  vol.~2, no. 2-4, pp. 241--266, Dec. 2008.

\bibitem{Francesca2014}
G.~Francesca, M.~Brambilla, A.~Brutschy, V.~Trianni, and M.~Birattari,
  ``Automode: A novel approach to the automatic design of control software for
  robot swarms,'' \emph{Swarm Intelligence}, vol.~8, no.~2, pp. 89--112, Jun
  2014.

\bibitem{Dallalibera2011}
F.~DallaLibera, S.~Ikemoto, T.~Minato, H.~Ishiguro, E.~Menegatti, and
  E.~Pagello, ``Biologically inspired mobile robot control robust to hardware
  failures and sensor noise,'' in \emph{RoboCup 2010: Robot Soccer World Cup
  XIV}, J.~Ruiz-del Solar, E.~Chown, and P.~G. Pl{\"o}ger, Eds.\hskip 1em plus
  0.5em minus 0.4em\relax Berlin, Heidelberg: Springer Berlin Heidelberg, 2011,
  pp. 218--229.

\bibitem{Claudi2014}
A.~{Claudi}, D.~{Accattoli}, P.~{Sernani}, P.~{Calvaresi}, and A.~F. {Dragoni},
  ``A noise-robust obstacle detection algorithm for mobile robots using active
  3d sensors,'' in \emph{Proceedings ELMAR-2014}, 2014, pp. 1--4.

\bibitem{Turgut2008}
A.~E. Turgut, H.~{\c{C}}elikkanat, F.~G{\"o}k{\c{c}}e, and E.~{\c{S}}ahin,
  ``Self-organized flocking in mobile robot swarms,'' \emph{Swarm
  Intelligence}, vol.~2, no.~2, pp. 97--120, Dec 2008.

\bibitem{Harwell2020a}
J.~Harwell, L.~Lowmanstone, and M.~Gini, ``Demystifying emergent intelligence
  and its effect on performance in large robot swarms,'' in \emph{Proc.
  Autonomous Agents and Multi-agent Systems (AAMAS)}, May 2020, pp. 474--482.

\bibitem{Winfield2005}
A.~F.~T. Winfield, C.~J. Harper, and J.~Nembrini, ``Towards dependable swarms
  and a new discipline of swarm engineering,'' in \emph{Swarm Robotics. SR
  2004. Lecture Notes in Computer Science}.\hskip 1em plus 0.5em minus
  0.4em\relax Springer, 2005, vol. LNCS 3342, pp. 126--142.

\bibitem{Brambilla2013a}
M.~Brambilla, E.~Ferrante, M.~Birattari, and M.~Dorigo, ``{Swarm robotics: A
  review from the swarm engineering perspective},'' \emph{Swarm Intelligence},
  vol.~7, no.~1, pp. 1--41, 2013.

\bibitem{Campo2007}
A.~Campo and M.~Dorigo, ``Efficient multi-foraging in swarm robotics,'' in
  \emph{Advances in Artificial Life, LNAI 4648}.\hskip 1em plus 0.5em minus
  0.4em\relax Springer, 2007.

\bibitem{Correll2008}
N.~Correll, ``Parameter estimation and optimal control of swarm-robotic
  systems: A case study in distributed task allocation,'' in \emph{Proc. IEEE
  Int'l Conf. on Robotics and Automation}, May 2008, pp. 3302--3307.

\bibitem{Moarref2018}
S.~Moarref and H.~Kress-Gazit, ``Reactive synthesis for robotic swarms,'' in
  \emph{Formal Modeling and Analysis of Timed Systems}, D.~N. Jansen and
  P.~Prabhakar, Eds.\hskip 1em plus 0.5em minus 0.4em\relax Cham: Springer
  International Publishing, 2018, pp. 71--87.

\bibitem{Talamali2020}
M.~S. Talamali, T.~Bose, M.~Haire, X.~Xu, J.~A.~R. Marshall, and A.~Reina,
  ``Sophisticated collective foraging with minimalist agents: a swarm robotics
  test,'' \emph{Swarm Intelligence}, vol.~14, no.~1, pp. 25--56, Mar 2020.

\bibitem{Berman2007}
S.~Berman, {\'{A}}.~Hal{\'{a}}sz, V.~Kumar, and S.~Pratt, ``Algorithms for the
  analysis and synthesis of a bio-inspired swarm robotic system,'' in
  \emph{Swarm Robotics}.\hskip 1em plus 0.5em minus 0.4em\relax Springer-Verlag
  Berlin Heidelberg, 2007, vol. LNCS 4433, pp. 56--70.

\bibitem{Lerman2001}
K.~Lerman, A.~Galstyan, A.~Martinoli, and A.~Ijspeert, ``{A macroscopic
  analytical model of collaboration in distributed robotic systems},''
  \emph{Artificial Life}, vol.~7, no.~4, pp. 375--393, 2001.

\bibitem{Ligot2020}
A.~Ligot, K.~Hasselmann, and M.~Birattari, ``Automode-arlequin: Neural networks
  as behavioral modules for the automatic design of probabilistic finite-state
  machines,'' in \emph{Swarm Intelligence}, M.~Dorigo, T.~St{\"u}tzle, M.~J.
  Blesa, C.~Blum, H.~Hamann, M.~K. Heinrich, and V.~Strobel, Eds.\hskip 1em
  plus 0.5em minus 0.4em\relax Cham: Springer International Publishing, 2020,
  pp. 271--281.

\bibitem{Hogg2020}
E.~Hogg, S.~Hauert, D.~Harvey, and A.~Richards, ``Evolving behaviour trees for
  supervisory control of robot swarms,'' \emph{Artificial Life and Robotics},
  vol.~25, no.~4, pp. 569--577, Nov 2020.

\bibitem{Panagou2020}
D.~{Panagou}, M.~{Turpin}, and V.~{Kumar}, ``Decentralized goal assignment and
  safe trajectory generation in multirobot networks via multiple lyapunov
  functions,'' \emph{IEEE Transactions on Automatic Control}, vol.~65, no.~8,
  pp. 3365--3380, 2020.

\bibitem{Glotfelter2019}
P.~{Glotfelter}, I.~{Buckley}, and M.~{Egerstedt}, ``Hybrid nonsmooth barrier
  functions with applications to provably safe and composable collision
  avoidance for robotic systems,'' \emph{IEEE Robotics and Automation Letters},
  vol.~4, no.~2, pp. 1303--1310, 2019.

\bibitem{Ames2021}
A.~D. {Ames}, G.~{Notomista}, Y.~{Wardi}, and M.~{Egerstedt}, ``Integral
  control barrier functions for dynamically defined control laws,'' \emph{IEEE
  Control Systems Letters}, vol.~5, no.~3, pp. 887--892, 2021.

\bibitem{Hsieh2013}
T.~W. {Hsieh, M. Ani, Mather}, ``{Robustness in the Presence of Task
  Differentiation in Robot Ensembles},'' \emph{Redundancy in Robot Manipulators
  and Multi-Robot Systems}, pp. 93--108, 2013.

\bibitem{Baumeister2020}
T.~Baumeister, B.~Finkbeiner, and H.~Torfah, ``Explainable reactive
  synthesis,'' in \emph{Automated Technology for Verification and Analysis},
  D.~V. Hung and O.~Sokolsky, Eds.\hskip 1em plus 0.5em minus 0.4em\relax Cham:
  Springer International Publishing, 2020, pp. 413--428.

\bibitem{Pacheck2020}
A.~{Pacheck}, S.~{Moarref}, and H.~{Kress-Gazit}, ``Finding missing skills for
  high-level behaviors,'' in \emph{2020 IEEE International Conference on
  Robotics and Automation (ICRA)}, 2020, pp. 10\,335--10\,341.

\bibitem{Birattari2019}
M.~Birattari \emph{et~al.}, ``Automatic off-line design of robot swarms: A
  manifesto,'' \emph{Frontiers in Robotics and AI}, vol.~6, p.~59, 2019.

\bibitem{Bjerknes2013}
J.~D. Bjerknes and A.~F.~T. Winfield, \emph{On Fault Tolerance and Scalability
  of Swarm Robotic Systems}.\hskip 1em plus 0.5em minus 0.4em\relax Berlin,
  Heidelberg: Springer Berlin Heidelberg, 2013, pp. 431--444.

\bibitem{Rubenstein2014}
M.~Rubenstein, C.~Ahler, N.~Hoff, A.~Cabrera, and R.~Nagpal, ``Kilobot: A low
  cost robot with scalable operations designed for collective behaviors,''
  \emph{Robotics and Autonomous Systems}, vol.~62, no.~7, pp. 966--975, 2014,
  reconfigurable Modular Robotics.

\bibitem{Harwell2018}
J.~Harwell and M.~Gini, ``Broadening applicability of swarm-robotic foraging
  through constraint relaxation,'' \emph{IEEE Int'l Conf. on Simulation,
  Modeling, and Programming for Autonomous Robots (SIMPAR)}, pp. 116--122, May
  2018.

\bibitem{Hamann2008}
H.~Hamann and H.~W{\"{o}}rn, ``{A framework of space-time continuous models for
  algorithm design in swarm robotics},'' \emph{Swarm Intelligence}, vol.~2, no.
  2-4, pp. 209--239, 2008.

\bibitem{George2005}
J.-P. Georg{\'{e}} and M.-P. Gleizes, ``{Experiments in Emergent Programming
  Using Self-organizing Multi-agent Systems},'' \emph{Multi-Agent Systems and
  Applications IV}, vol. 3690, pp. 450--459, 2005.

\bibitem{Tarapore2020}
D.~Tarapore, R.~Groß, and K.-P. Zauner, ``Sparse robot swarms: Moving swarms
  to real-world applications,'' \emph{Frontiers in Robotics and AI}, vol.~7,
  p.~83, 2020.

\bibitem{Frison2010}
M.~Frison, N.~L. Tran, N.~Baiboun, A.~Brutschy, G.~Pini, A.~Roli, M.~Dorigo,
  and M.~Birattari, ``Self-organized task partitioning in a swarm of robots,''
  in \emph{Swarm Intelligence}.\hskip 1em plus 0.5em minus 0.4em\relax
  Springer, Berlin, Heidelberg, 2010, vol. LNCS 6234, pp. 287--298.

\bibitem{Li2016}
H.~Li, T.~Wang, H.~Wei, and C.~Meng, ``Response strategy to environmental cues
  for modular robots with self-asssembly from swarm to articulated robots,''
  \emph{Journal of Intelligent and Robotic Systems: Theory and Applications},
  vol.~81, no. 3-4, pp. 359--376, 2016.

\bibitem{Beni2005}
G.~Beni, ``Order by disordered action in swarms,'' in \emph{LNCS 3342}.\hskip
  1em plus 0.5em minus 0.4em\relax Springer, 2005.

\bibitem{Bonabeau1999}
E.~Bonabeau, M.~Dorigo, D.~d. R. D.~F. Marco, G.~Theraulaz, G.~Th{\'e}raulaz
  \emph{et~al.}, \emph{Swarm intelligence: from natural to artificial
  systems}.\hskip 1em plus 0.5em minus 0.4em\relax Oxford University Press,
  1999, no.~1.

\bibitem{Correll2006}
N.~Correll and A.~Martinoli, \emph{{Towards optimal control of self-organized
  robotic inspection systems}}.\hskip 1em plus 0.5em minus 0.4em\relax IFAC,
  2006, vol.~8, no. PART 1.

\bibitem{Payton2001}
D.~Payton, M.~Daily, R.~Estowski, M.~Howard, and C.~Lee, ``{Pheromone
  robotics},'' \emph{Autonomous Robots}, vol.~11, no.~3, pp. 319--324, 2001.

\bibitem{Beni1993}
G.~Beni and J.~Wang, ``Swarm intelligence in cellular robotic systems,'' in
  \emph{Robots and Biological Systems: Towards a New Bionics?}, P.~Dario,
  G.~Sandini, and P.~Aebischer, Eds.\hskip 1em plus 0.5em minus 0.4em\relax
  Berlin, Heidelberg: Springer Berlin Heidelberg, 1993, pp. 703--712.

\bibitem{Sugawara1997}
K.~Sugawara and M.~Sano, ``{Cooperative acceleration of task performance:
  Foraging behavior of interacting multi-robots system},'' \emph{Physica D:
  Nonlinear Phenomena}, vol. 100, no. 3-4, pp. 343--354, 1997.

\bibitem{Petersen2011}
K.~Petersen, R.~Nagpal, and J.~Werfel, ``{TERMES}: An autonomous robotic system
  for three-dimensional collective construction,'' in \emph{Robotics: Science
  and Systems (RSS)}, 2011.

\bibitem{Nave2020}
G.~K. Nave, N.~T. Mitchell, J.~A. Chan~Dick, T.~Schuessler, J.~A. Lagarrigue,
  and O.~Peleg, ``Attraction, dynamics, and phase transitions in fire ant
  tower-building,'' \emph{Frontiers in Robotics and AI}, vol.~7, p.~25, 2020.

\bibitem{Rosenfeld2006}
A.~Rosenfeld, G.~A. Kaminka, and S.~Kraus, ``A study of scalability properties
  in robotic teams,'' in \emph{Coordination of Large-Scale Multiagent Systems},
  2006, pp. 27--51.

\bibitem{Lopes2016a}
Y.~K. Lopes, S.~M. Trenkwalder, A.~B. Leal, T.~J. Dodd, and R.~Gro{\ss},
  ``{Supervisory control theory applied to swarm robotics},'' \emph{Swarm
  Intelligence}, vol.~10, no.~1, pp. 65--97, 2016.

\bibitem{Hamann2013}
H.~Hamann, ``Towards swarm calculus: urn models of collective decisions and
  universal properties of swarm performance,'' \emph{Swarm Intelligence},
  vol.~7, pp. 145--172, 2013.

\bibitem{Liu2009}
B.~Liu, T.~Chu, and L.~Wang, ``{Collective motion in non-reciprocal swarms},''
  \emph{Journal of Control Theory and Applications}, vol.~7, no.~2, pp.
  105--111, 2009.

\bibitem{Ferrante2015}
E.~Ferrante, A.~E. Turgut, E.~Du{\'{e}}{\~{n}}ez-Guzm{\'{a}}n, M.~Dorigo, and
  T.~Wenseleers, ``Evolution of self-organized task specialization in robot
  swarms,'' \emph{PLoS Computational Biology}, vol.~11, no.~8, 2015.

\bibitem{Pini2011a}
G.~Pini, A.~Brutschy, M.~Birattari, and M.~Dorigo, ``Task partitioning in
  swarms of robots: Reducing performance losses due to interference at shared
  resources,'' in \emph{Informatics in Control Automation and Robotics}, ser.
  LNEE 85.\hskip 1em plus 0.5em minus 0.4em\relax Springer, 2011, pp. 217--228.

\bibitem{Karp1990}
A.~H. Karp and H.~P. Flatt, ``Measuring parallel processor performance,''
  \emph{Commun. ACM}, vol.~33, no.~5, pp. 539--543, May 1990.

\bibitem{Duarte2016}
M.~Duarte, V.~Costa, J.~Gomes, T.~Rodrigues, F.~Silva, S.~M. Oliveira, and
  A.~L. Christensen, ``Evolution of collective behaviors for a real swarm of
  aquatic surface robots,'' \emph{PLoS ONE}, vol.~11, no.~3, pp. 1--25, 2016.

\bibitem{Ferrante2013a}
E.~Ferrante, E.~Duenez-Guzman, A.~E. Turgut, and T.~Wenseleers, ``{GESwarm}:
  grammatical evolution for the automatic synthesis of collective behaviors in
  swarm robotics,'' \emph{Proc. Conf. on Genetic and Evolutionary Computation
  (GECCO)}, pp. 17--24, 2013.

\bibitem{Jekel2018}
C.~F. Jekel, G.~Venter, M.~P. Venter, N.~Stander, and R.~T. Haftka,
  ``{Similarity measures for identifying material parameters from hysteresis
  loops using inverse analysis},'' \emph{International Journal of Material
  Forming}, pp. 1--24, 2018.

\bibitem{Tarapore2017}
D.~Tarapore, A.~L. Christensen, and J.~Timmis, ``Generic, scalable and
  decentralized fault detection for robot swarms,'' \emph{PLOS ONE}, vol.~12,
  no.~8, pp. 1--29, 08 2017.

\bibitem{GuerreroBonilla2020b}
L.~{Guerrero-Bonilla}, D.~{Saldaña}, and V.~{Kumar}, ``Dense r-robust
  formations on lattices,'' in \emph{2020 IEEE International Conference on
  Robotics and Automation (ICRA)}, 2020, pp. 6633--6639.

\bibitem{Usevitch2019}
J.~{Usevitch} and D.~{Panagou}, ``Determining r-robustness of digraphs using
  mixed integer linear programming,'' in \emph{2019 American Control Conference
  (ACC)}, 2019, pp. 2257--2263.

\bibitem{Zhang2008}
Y.~Zhang, F.~Bastani, I.~L. Yen, F.~Jicheng, and I.~R. Chen, ``{Availability
  analysis of robotic swarm systems},'' \emph{Proc. 14th IEEE Pacific Rim Int'l
  Symposium on Dependable Computing, PRDC 2008}, pp. 331--338, 2008.

\bibitem{Seda2017}
M.~{\v{S}}eda, J.~{\v{S}}edov{\'{a}}, and M.~Hork{\'{y}}, ``Models and
  simulations of queueing systems,'' in \emph{Recent Advances in Soft
  Computing. Proc. 22nd International Conference on Soft Computing (MENDEL
  2016)}, R.~Matou{\v{s}}ek, Ed.\hskip 1em plus 0.5em minus 0.4em\relax
  Springer, 2017, vol. 576.

\bibitem{Rouff2007a}
C.~Rouff, ``Intelligence in future {NASA} swarm-based missions,'' in
  \emph{Regarding the Intelligence in Distributed Intelligent Systems, Papers
  from the 2007 AAAI Fall Symposium}, T.~Finin, L.~Kagal, E.~F. Kendall, J.~H.
  Li, M.~Lyell, and W.~Truszkowski, Eds., 2007, vol. AAAI, FS-07-06, pp.
  112--115.

\bibitem{Gazi2004}
V.~Gazi and K.~M. Passino, ``{Stability Analysis of Social Foraging Swarms},''
  \emph{IEEE Transactions on Systems, Man, and Cybernetics, Part B:
  Cybernetics}, vol.~34, no.~1, pp. 539--557, 2004.

\bibitem{Renshaw1981}
E.~Renshaw and R.~Henderson, ``The correlated random walk,'' \emph{Journal of
  Applied Probability}, vol.~18, no.~2, pp. 403--414, 1981.

\bibitem{Pinciroli2012}
C.~Pinciroli \emph{et~al.}, ``Argos: a modular, parallel, multi-engine
  simulator for multi-robot systems,'' \emph{Swarm Intelligence}, vol.~6, pp.
  271--295, 12 2012.

\bibitem{Dorigo2005c}
M.~Dorigo, ``Swarm-bot: An experiment in swarm robotics,'' in \emph{Proc. IEEE
  Swarm Intelligence Symposium}, 2005, pp. 199--207.

\bibitem{Hsieh2008}
M.~A. Hsieh, {\'A}.~Hal{\'a}sz, S.~Berman, and V.~Kumar, ``Biologically
  inspired redistribution of a swarm of robots among multiple sites,''
  \emph{Swarm Intelligence}, vol.~2, no.~2, pp. 121--141, Dec 2008.

\end{thebibliography}

\end{document}